\documentclass[letterpaper, 10 pt, journal, twoside]{IEEEtran}
\usepackage{amsmath,amsfonts}
\usepackage{algorithmic}
\usepackage{algorithm}
\usepackage{array}
\usepackage[caption=false,font=normalsize,labelfont=sf,textfont=sf]{subfig}
\usepackage{textcomp}
\usepackage{stfloats}
\usepackage{url}
\usepackage{verbatim}
\usepackage{graphicx}
\usepackage{booktabs,multirow}
\usepackage{cite}
\usepackage{makecell}
\usepackage{threeparttable}
\begin{document}

% Paper headers
\markboth{IEEE Robotics and Automation Letters. Preprint Version. Accepted March 2023}
{Zhuang \MakeLowercase{\textit{et al.}}: 4D iRIOM: 4D Imaging Radar Inertial Odometry and Mapping}
%Use only for final RAL version

\title{4D iRIOM: 4D Imaging Radar Inertial\\Odometry and Mapping}

\author{Yuan Zhuang$^{1}$,~\IEEEmembership{Senior Member, IEEE,} Binliang Wang$^{1}$, \\ Jianzhu Huai$^{1*}$, ~\IEEEmembership{Member, IEEE,} Miao Li$^{1}$, ~\IEEEmembership{Senior Member, IEEE}
\thanks{Manuscript received: December 6, 2022; Revised: February 19, 2023; Accepted: March 20, 2023.
This paper was recommended for publication by Editor Javier Civera upon evaluation of the Associate Editor and Reviewers’ comments.
This work was supported by the international research program of the national key R\&D plan of the China Ministry of Science and Technology (2022YFE0139300).}
\thanks{$^{1}$Authors are with State Key Lab of Info Engineering in Surveying, Mapping, and Remote Sensing, Wuhan University, China}
\thanks{$^*$Corresponding author email: jianzhu.huai@whu.edu.cn}
\thanks{Digital Object Identifier (DOI): see top of this page.}}
% Use only for final RAL version.


\maketitle

\begin{abstract}
Millimeter wave radar can measure distances, directions, and Doppler velocity for objects in harsh conditions such as fog.
The 4D imaging radar with both vertical and horizontal data resembling an image can also measure objects' height.
Previous studies have used 3D radars for ego-motion estimation.
But few methods leveraged the rich data of imaging radars, and they usually omitted the mapping aspect, thus leading to inferior odometry accuracy.
This paper presents a real-time imaging radar inertial odometry and mapping method, iRIOM, based on the submap concept.
To deal with moving objects and multipath reflections, we use the graduated non-convexity method to robustly and efficiently estimate ego-velocity from a single scan.
To measure the agreement between sparse non-repetitive radar scan points and submap points, the distribution-to-multi-distribution distance for matches is adopted.
The ego-velocity, scan-to-submap matches are fused with the 6D inertial data by an iterative extended Kalman filter to get the platform's 3D position and orientation.
A loop closure module is also developed to curb the odometry module's drift.
To our knowledge, iRIOM based on the two modules is the first 4D radar inertial SLAM system.
On our and third-party data, we show iRIOM's favorable odometry accuracy and mapping consistency against the FastLIO-SLAM and the EKFRIO.
Also, the ablation study reveal the benefit of inertial data versus the constant velocity model, and scan-to-submap matching versus scan-to-scan matching.
\end{abstract}

\begin{IEEEkeywords}
4D imaging radar, radar inertial odometry and mapping, scan-to-submap, graduated non-convexity.
\end{IEEEkeywords}

\section{Introduction}
\IEEEPARstart{A}{utonomous} vehicles and robotics are developing rapidly recently. 
Positioning and navigation technology based on sensor fusion has been a popular research field. 
Currently, most platforms use lidars and cameras as primary sensors.
These sensors have good performance in good weather, but struggle to work in degraded environments such as rain and fog \cite{dickmann2016automotive}.
The millimeter wave radar typically has a longer wavelength and a larger detection range, and is barely susceptible to small particles such as fog and rain \cite{brooker2001millimetre}.
Moreover, it can provide Doppler velocity observations, leading to ego-velocity from a single radar scan \cite{kellner2013instantaneous}. 

Compared to 3D radars, the 4D imaging millimeter wave radar greatly improves the resolution in the vertical direction, 
thus providing four types of information: distance, azimuth, height, and velocity.
It allows better obstacle detection, ego-motion estimation, and path planning \cite{sun20214d}.
However, a 4D millimeter wave radar obtains scans of sparse point clouds which are lower in spatial resolution than camera images or lidar scans at the moment.
Moreover, radar scans suffer from multipath effects, harmonics and other noises, making it difficult to achieve reliable data association \cite{doer2021yaw}.
Thus, it is a challenging task to attain precise 3D localization and mapping by a 4D millimeter wave radar.

\begin{figure}[t]
	\centering
	\includegraphics[width=8cm]{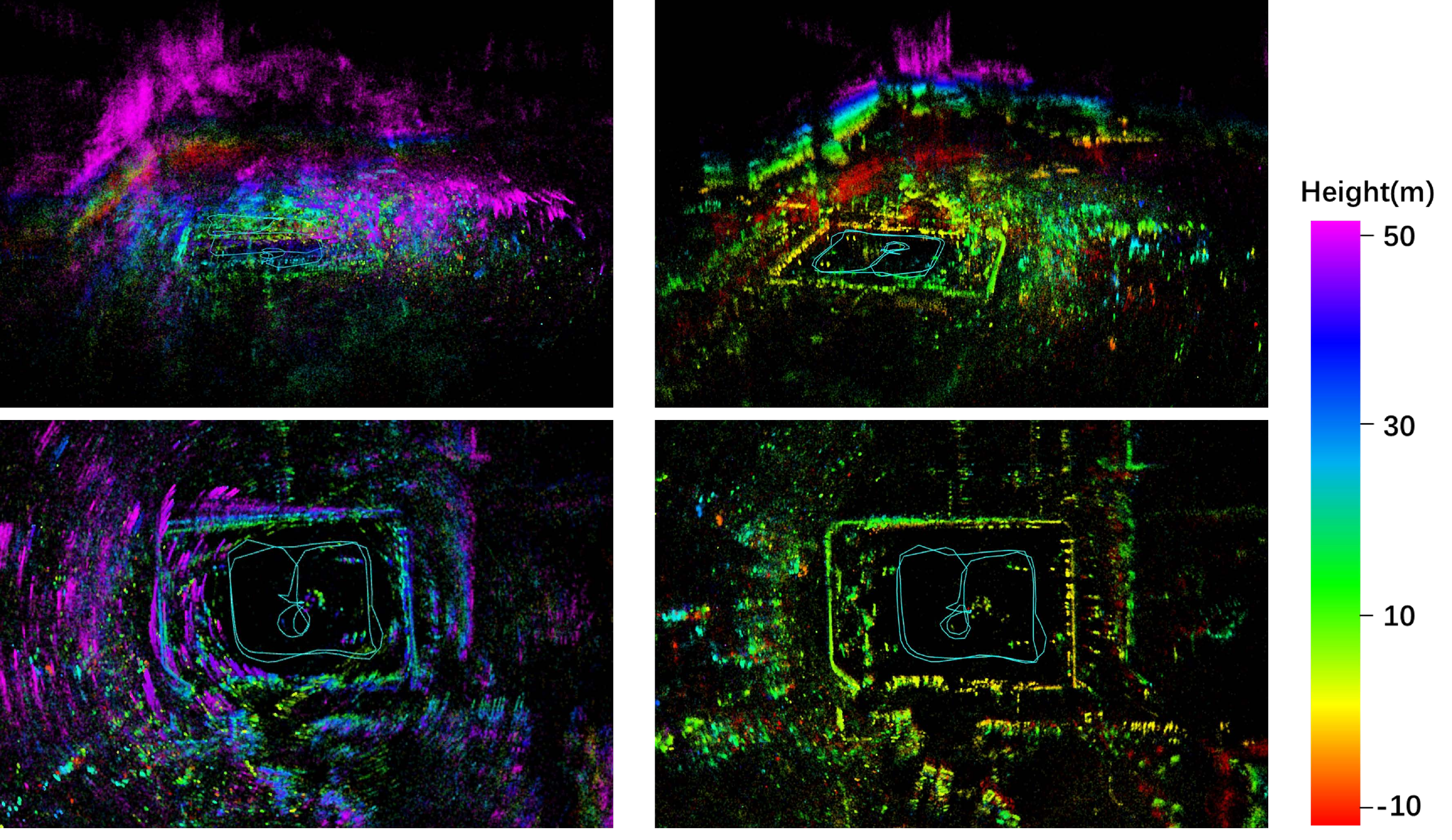}
	\caption{Mapping results by iRIOM on data of a basketball court.
Radar ego-velocity and IMU data are used for the left plots, and radar point matches are further added for the right plots.
The blue curves are the estimated trajectories.
Map points are colored by the elevation.}
	\label{fig:mapping-data1}
\end{figure}

We propose an imaging radar inertial odometry and mapping method (iRIOM) using radar scans of point clouds and 6D IMU (inertial measurement unit) data.
The iRIOM fuses ego-velocity from radar scans and scan-to-submap matches with IMU data to get accurate odometry and consistent mapping.
Our contributions are threefold:

1. \textbf{Novel 4D radar measurement procurement}:
To estimate ego-velocity from a noisy radar scan, we use the robust graduated non-convexity (GNC) rather than the common RANSAC \cite{doer2020ekf}.
To match sparse radar data,
point matches expressed as distribution to multi-distribution constraints are found by matching the current scan to a submap built by the mapping module, 
instead of scan-to-scan matching \cite{michalczyktightly}.

2. \textbf{Complete 4D radar-inertial SLAM system}: Based on an iterative extended Kalman filter (EKF),
we fuse IMU data with ego-velocity and point match constraints, forming an odometry method.
To handle place revisits, we adopt an existing method, scancontext \cite{kim2018scan}, to recognize previous radar scans, and then close loops in the pose graph, 
leading to the first complete 4D radar-inertial simultaneous localization and mapping (SLAM) system to our knowledge.

3. \textbf{Extensive experiments}: 
The tests of the SLAM system on indoor and outdoor sequences of our dataset and the ColoRadar dataset \cite{kramer2022coloradar},
show that the system's localization accuracy surpasses existing radar-inertial odometry approaches and approaches lidar-inertial SLAM approaches.
Moreover, the ablation studies reveal the benefits of IMU data, the ego-velocity measurements, and the scan-to-submap point matches.

The rest of this paper is organized as follows. 
Section II reviews the related work on radar odometry and radar inertial odometry. 
Section III describes our proposed 4D radar inertial odometry method. 
Then, we experimentally validate the proposed method and analyze the results in Section IV.
In Section V, we summarize the paper and draw conclusions.

\section{RELATED WORK}
We briefly review related work on 3D radar odometry, 4D radar inertial odometry, and radar-based place recognition.
\subsection{3D Radar Odometry}
Conventional millimeter wave radars can only estimate planar position based on distance and azimuth data, 
and cannot get object heights due to the low elevation resolution. 
Thus, most radar odometry methods convert radar data to planar images or 2D point clouds, and then estimate odometry using image or point cloud matching methods. 
Based on the data association, these odometry methods can be grouped into feature-based and direct methods.

The feature-based approach first extracts key points and feature descriptors, 
and then associates key points by nearest neighbor searching in the feature space or graph matching without descriptors.
Cen et al. \cite{cen2018precise}, \cite{cen2019radar} proposed two radar key point detection methods, 
and showed that the two methods can suppress multipath noise and improve robustness of sequential radar scan matching.
Aldera et al. \cite{aldera2019fast} trained a focus of attention policy by a weakly supervised learning approach to down-sample the point clouds as preprocessing, 
reducing the time consumption while ensuring accuracy of the radar odometer. 
Barnes et al. \cite{barnes2020under} proposed a self-supervised radar key point detection and 
feature description network to generate good key point locations and descriptors for scan matching, 
which substantially improved accuracy of radar odometry. 
Hong et al. \cite{hong2020radarslam} used classical feature point detection and description algorithms (e.g., SURF) to 
associate adjacent scans to achieve pose estimation as the front end of a radar SLAM system, 
and used loop detection for global pose optimization.
Burnett et al. \cite{burnett2021we} used the key point extraction and description algorithm proposed in \cite{cen2019radar} to 
build a radar SLAM system and considered motion distortion in the rotating radar data. 

The direct methods estimate the relative pose by directly registering radar scans.
Park et al. \cite{park2020pharao} proposed a direct radar odometry method by applying the Fourier-Melling transform (FMT) to 
estimate rotations and translations between sequential radar scans. 
In \cite{barnes2019masking}, Barnes et al. proposed a fully microscopic, correlation-based end-to-end radar odometry, 
where a mask network was trained to remove radar data noise, and the fast Fourier transform (FFT) matching method was used to obtain the relative pose.

\subsection{4D Radar Inertial Odometry}
Compared to 3D radars, 4D imaging radars significantly improve the elevation resolution, though its point clouds are still very sparse ($\sim400$ points/scan).
Meanwhile, the search space of 4D imaging radar point cloud matching changes from 2D to 3D, and the matching process is affected by various types of noise, 
thus, it is difficult to achieve stable, continuous, and high-precision scan matching.

To make the odometry more robust and mitigate trajectory drift for missing data and wrong matches in scan-to-scan matching, 
several works used inertial data as continual observations along with scan-to-scan matching.
Almalioglu et al. \cite{almalioglu2020milli} coupled the radar inter-frame poses obtained from the normal distribution transform and IMU data by an unscented Kalman filter, 
and used a long short-term memory motion model to estimate 3D ego-motion. 
Lu et al. \cite{lu2020milliego} proposed a deep neural network based radar inertial odometer, 
which encoded radar data by a convolutional neural network and IMU data by a recurrent neural network, 
and achieved real-time pose estimation by a multi-modal sensor fusion network.

Since 4D radar can estimate the 3D ego-velocity from a single scan, many studies coupled IMU measurements with radar single-scan velocity estimates. 
Kramer et al. \cite{kramer2020radar} proposed a robot velocity estimation method based on sliding window optimization to fuse inertial data with radar ego-velocity, estimating the robot's 3D motion in real time.
Doer et al. \cite{doer2020ekf} proposed a radar inertial odometry method based on the extended Kalman filter (EKF)
for 3D localization by fusing IMU data with radar ego-velocity estimated by a RANSAC scheme.
Moreover, altimeter data were used to eliminate drift in altitude.
Subsequently, they \cite{doer2021yaw} introduced online calibration and the Manhattan world assumption as a yaw angle aid for odometry.

%Other works considered that both scan matching and IMU propagation can lead to positioning error and trajectory drift. 
Overall, scan matching can give many point constraints, but is sensitive to measurement noise and matching errors \cite{li2021real}, and may break down in areas of sparse radar points.
The ego-velocity derived from a radar scan is not affected by drift, 
but is largely unable to correct vertical drift (as shown in Fig.~\ref{fig:mapping-data1}).
Combining their advantages, we propose a radar inertial odometry method that fuses and single-scan ego-velocity and scan matching which is enhanced by the scan-to-submap concept.

To remedy drift of an odometry method, a loop closure module is needed.
For radar-based systems, the current methods for place recognition can be divided into 
deep learning-based methods, e.g., \cite{wang2021radarloc}, and traditional geometric methods, e.g., \cite{park2020pharao}, \cite{kim2018scan}.
Learning-based methods can achieve good results for some scenes, but their computation cost and generality are a concern.
Thus, we adopt a geometric method, scancontext \cite{kim2018scan}, for loop detection.

\section{METHODOLOGY}

\subsection{Framework Overview}
\begin{figure}[thpb]
	\centering
	\includegraphics[width=8.5cm]{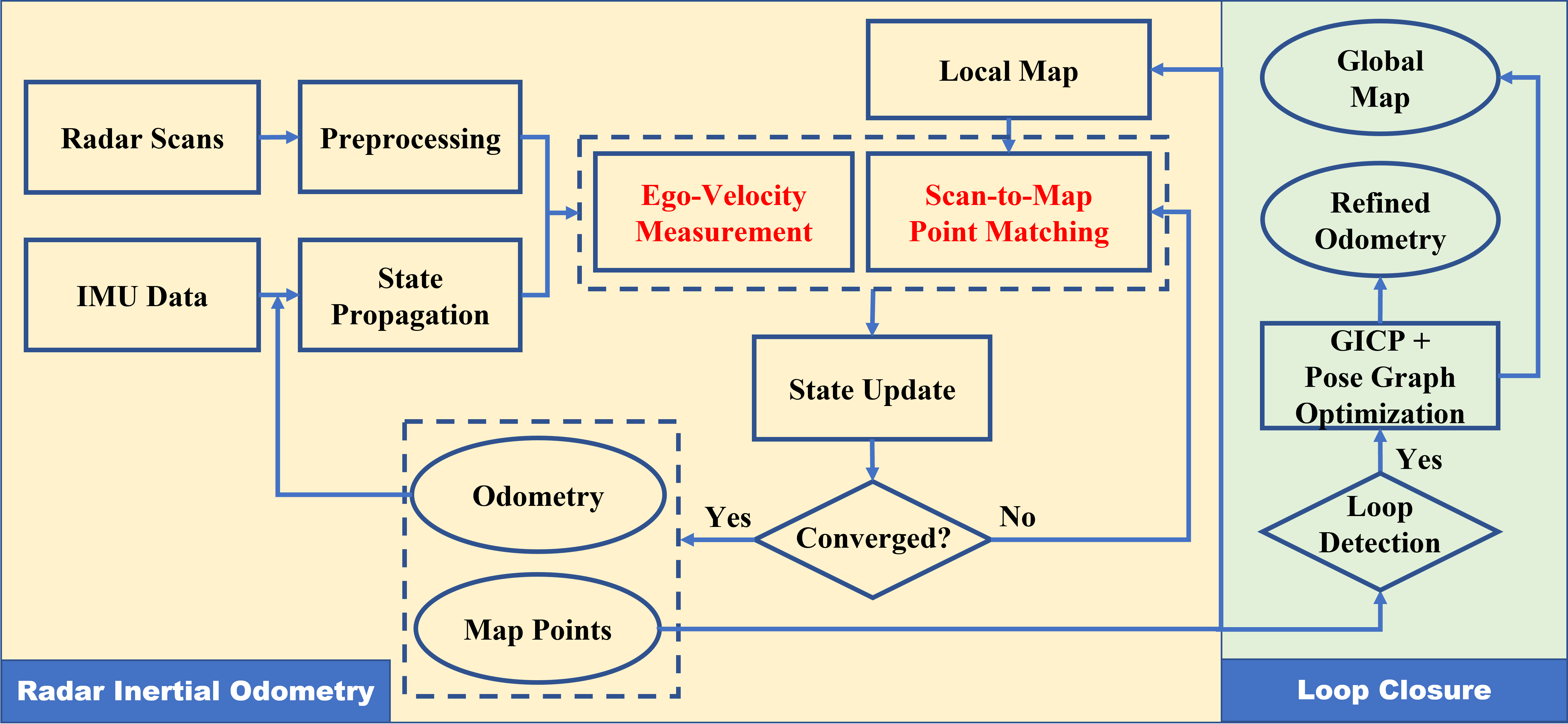}
	\caption{System overview of iRIOM.}
	\label{fig:workflow}
\end{figure}

The proposed iRIOM system is shown in Fig.~\ref{fig:workflow}.
The inputs are time-aligned radar scans of point clouds and 6D IMU data. 
To deal with noise points caused by multipath and scattering,
we first filter out these points with some spatial statistics in the pre-processing. 
Second, we propagate the system state (position, attitude, velocity, etc.) by IMU data,
estimate the ego-velocity from the incoming radar scan,
and obtain point matches by scan-to-submap matching.
The ego-velocity and point matches weighted by their distribution covariances are used to update the system state in an iterated Kalman filter.
Third, the radar scan is transformed into the global coordinate system by the estimated pose to update the local submap and the global map.
Finally, loops are detected by scancontext \cite{kim2018scan} which works well for range data captured without sensor tilting.
We only adjusted its descriptor distance threshold for 4D radars.
Loop constraints are computed by the generalized ICP \cite{koide2021voxelized} and used in pose graph optimization along with relative pose constraints, obtaining consistent poses and maps.

\subsection{Radar Data Preprocessing}
Compared to sensors such as lidars and cameras, there is a greater proportion of invalid points in the scans by radar, 
including interference points and noise points due to multipath reflections and speckles, 
which may degrade mapping and positioning.
Thus, we pre-process the radar data to remove such points. 

In order to achieve an optimal denoising effect while maintaining sufficient radar point observations, 
we first remove noise points by relaxation filtering which considers spatial distributions \cite{balta2018fast}.
The rules are defined by
\begin{equation}
	\begin{aligned}
		\mathbf{P} = &\left\{ \mathbf{p}_{i} \middle| Num\left( \mathbf{p}_{j} \right) > N_{th}~and~\left\| {\mathbf{p}_{i} - \mathbf{p}_{j}} \right\| < D_{th} \right\} \\
		\mathbf{P} = &\left\{ \mathbf{p}_{i} \middle| Std\left( \mathbf{p}_{j} \right) < \sigma_{th}~and~\left\| {\mathbf{p}_{i} - \mathbf{p}_{j}} \right\| < D_{th} \right\}
	\end{aligned}
\end{equation}
where $\mathbf{P}$ is the set of inlier points, $\mathbf{p}_{j}$ is a neighboring point of $\mathbf{p}_{i}$ within the neighborhood radius $D_{th}$, 
$Num\left( \mathbf{p}_{j} \right)$ is the number of neighborhood points,
$Std\left( \mathbf{p}_{j} \right)$ is the statistical standard deviation of the neighboring points, 
and $N_{th}$ and $\sigma_{th}$ are their corresponding thresholds.

Points due to moving objects are found in the ego-velocity estimation step which estimates the velocity of the platform relative to the static world.
For a static world point at $\boldsymbol{\rho}$ in the radar coordinate system $\{R\}$, its 1D Doppler velocity measurement $v^d$ relates to the ego-velocity of the radar in $\{R\}$, ${}^{R}\mathbf{v}_{m}$, by 
$v^d = {\boldsymbol{\rho}^T} \cdot {}^{R}\mathbf{v}_{m} /{||\boldsymbol{\rho}||}$.

Estimation of the ego-velocity with Doppler velocity measurements in a radar scan is often 
done by a least squares method, e.g., \cite{kramer2020radar,doer2020ekf,doer2021yaw}.
To deal with moving points, iterative methods with varying weights or RANSAC methods can be used.
Considering that a radar scan is sparse (typical \textless 500 points) and may contain many moving points,
we adopt the GNC \cite{yangGraduatedNonconvexityRobust2020} to estimate ego-velocity.
As a non-minimal solver, the GNC is typically more robust than RANSAC \cite{yangGraduatedNonconvexityRobust2020}.
Also, it is more efficient semi-definite programming \cite{yangCertifiablyOptimalOutlierrobust2022} which provides an optimality certifiable but is unsuited to problems with \textgreater100 observations.
With GNC, the ego-velocity estimation is solved by repeating three steps:
1) estimate $^{R}\mathbf{v}_{m}$ from the linear least squares problem
$\underset{^{R}\mathbf{v}_{m}}{\operatorname{min}} \sum_{i=1}^{n} w_i \left\|v_{i}^{d}-{\boldsymbol{\rho}_{i}^T} \cdot {}^{R}\mathbf{v}_{m}/{||\boldsymbol{\rho}_{i}||} \right\|^2$;
2) estimate weight $w_i$ as
$w_i$ = $\left(\mu \bar{c}^2 / (\hat{r}_i^2 + \mu \bar{c}^2)\right)^2$ where
$\hat{r}_i \dot{=} v_{i}^{d}-{\boldsymbol{\rho}_{i}^T} \cdot {}^{R}\mathbf{v}_{m}/{||\boldsymbol{\rho}_{i}||} $;
3) update $\mu\leftarrow \mu / 1.4$.
At the start, all weights are initialized to one, $\mu = 4 r_{max}^2 / \bar{c}^2$,
$\bar{c} = 2\sigma_r$ where $r_{max}^2$ is the maximum squared residual error and $\sigma_r$ is the accuracy of Doppler velocity.
The loop terminates until $\mu < 1$.
Finally, we get ${}^{R}\mathbf{v}_{m}^{~}$ noise covariance $\mathbf{R}_{vel}$ by inverting the information matrix in step 1).
Also, points with $r_i^2 < \bar{c}^2$ are taken as static points and input to the subsequent localization and mapping.

\subsection{IMU State Propagation}
We denote by $\mathbf{x}$ the state, 
$\mathbf{u}$ the control input (i.e., IMU data), and $\mathbf{w}$ the system noise,
\begin{equation}
	\begin{aligned}
		&\mathbf{x} = \left\lbrack {{}^{G}\mathbf{p}_{I}},{{}^{G}\mathbf{v}_{I}},{{}^{G}\mathbf{R}_{I}},\mathbf{b}_{a},\mathbf{b}_{\omega},{{}^{I}\mathbf{R}_{R}},{{}^{I}\mathbf{l}_{R}},{{}^{G}\mathbf{g}} \right\rbrack
		\\[1mm]
		&\mathbf{u} = \left\lbrack \mathbf{a}_{m},\boldsymbol{\omega}_{m} \right\rbrack
		\quad \mathbf{w} = \left\lbrack \mathbf{n}_{a},\mathbf{n}_{\omega},\mathbf{n}_{ba},\mathbf{n}_{b\omega}\right\rbrack
	\end{aligned}
\end{equation}
where $~{{}^{G}\mathbf{p}_{I}},{{}^{G}\mathbf{v}_{I}},{{}^{G}\mathbf{R}_{I}}$ are the position, velocity, and attitude of the IMU in the global coordinate system, resp.
The global frame is the IMU frame at the start of the SLAM system.
Since the global frame's $z$-axis may not coincide with the gravity direction, 
the gravity ${{}^{G}\mathbf{g}}$ is put in the state and its direction is estimated.
$\mathbf{b}_{a},\mathbf{b}_{\omega}$ are the accelerometer and gyro biases modeled as a random walk process with Gaussian noise $\mathbf{n}_{a}$,$\mathbf{n}_{\omega}$, and 
${{}^{I}\mathbf{R}_{R}},{{}^{I}\mathbf{l}_{R}}$ are the orientation and position of the radar in the IMU frame.
Thus, the extrinsic parameters can be estimated online as in \cite{doer2021yaw,xu2022fast} in case that they are inaccurate.
They may be locked by setting their initial covariance to zero.
$\mathbf{a}_{m},\boldsymbol{\omega}_{m}$ are the linear acceleration and angular velocity data with Gaussian process noises of $\mathbf{n}_{a},\mathbf{n}_{\omega}$, resp.

The discretized propagation model at $\mathbf{x} = \mathbf{x}_{i}$ with IMU data is given by
\begin{equation}
	\begin{aligned}
		\mathbf{x}_{i + 1}(1:3) &\approx \begin{bmatrix}
			{~{{}^{G}\mathbf{p}_{I_i}}+{}^{G}\mathbf{v}_{I_{i}}^{~}\mathrm{\Delta}t} \\
			{{{}^{G}\mathbf{v}_{I_i}}+\left({{}^{G}\mathbf{R}_{I_{i}}^{~}}\left( {\mathbf{a}_{m} - \mathbf{b}_{a_{i}}} \right) + {{{}^{G}\mathbf{g}}_{i}}^{~}\right)\mathrm{\Delta}t} \\
			{{^{G}\mathbf{R}_{I_{i}}} \cdot exp\left(\left(\boldsymbol{\omega}_{m} - \mathbf{b}_{\omega_{i}} \right)\mathrm{\Delta}t\right)}
		\end{bmatrix} \\
	     \mathbf{x}_{i + 1}(4:8) &\approx \mathbf{x}_{i}(4:8)
	\end{aligned}
\end{equation}
where $\mathrm{\Delta}t$ is the IMU sampling interval, and the indices 1:3 refer to the first three variables of $\mathbf{x}_{i+1}$.

We define the error state as in \cite{xu2022fast},
\begin{equation}
	\widetilde{\mathbf{x}} = \left\lbrack ~{{}^{G}{\widetilde{\mathbf{p}}}_{I}},
	{{}^{G}{\widetilde{\mathbf{v}}}_{I}},
	{{}^{G}{\widetilde{\mathbf{R}}}_{I}},
	{\widetilde{\mathbf{b}}}_{a},
	{\widetilde{\mathbf{b}}}_{\omega},
	{{}^{I}{\widetilde{\mathbf{R}}}_{L}},
	{{}^{I}{\widetilde{\mathbf{l}}}_{L}},
	{{}^{G}\widetilde{\mathbf{g}}} \right\rbrack.
\end{equation}
Propagation of the error state and its covariance  $\widehat{\mathbf{P}}$ is given by 
\begin{align}
		{\widetilde{\mathbf{x}}}_{i + 1} &= 
		\mathbf{x}_{i + 1} \ominus
		{\widehat{\mathbf{x}}}_{i + 1} 
		\approx \mathbf{F}_{\widetilde{\mathbf{x}}}{\widetilde{\mathbf{x}}}_{i} + \mathbf{F}_{\mathbf{w}}\mathbf{w}_{i} \\
	{\widehat{\mathbf{P}}}_{i + 1} &= \mathbf{F}_{\widetilde{\mathbf{x}}}{{\mathbf{P}}}_{i + 1}\mathbf{F}_{\widetilde{\mathbf{x}}}^{T} + \mathbf{F}_{\mathbf{w}}{{\mathbf{Q}}}_{i + 1}\mathbf{F}_{\mathbf{w}}^{T}
\end{align}
where ${\widehat{\mathbf{x}}}_{i + 1}$ is the state predicted by IMU propagation, $\ominus$ is the state minus operation, in which 
${^{G}\mathbf{R}_{I_{i + 1}}} \ominus {^{G}\widehat{\mathbf{R}}_{I_{i + 1}}} = \log({^{G}\mathbf{R}_{I_{i + 1}}}{^{G}\widehat{\mathbf{R}}_{I_{i + 1}}^{T}})$,
${^{I_{i + 1}}\mathbf{R}_{R_{i + 1}}} \ominus {^{I_{i + 1}}\widehat{\mathbf{R}}_{R_{i + 1}}}= \log({^{I_{i + 1}}\mathbf{R}_{R_{i + 1}}}{^{I_{i + 1}}\widehat{\mathbf{R}}_{R_{i + 1}}^{T}})$, 
and the others follow the common minus operation.
\begin{align}
		\mathbf{F}_{\widetilde{\mathbf{x}}} &= \begin{bmatrix}
			\mathbf{F}_\mathbf{A}  & \mathbf{F}_{\mathbf{B}} \\
			\mathbf{0}_{15\times9}  & {\mathbf{I}_{15\times15}} \\
		\end{bmatrix}\\
		\mathbf{F}_\mathbf{A} = \begin{bmatrix}
			\mathbf{I}  & {\mathbf{I}\mathrm{\Delta}t}  &\mathbf{0}\\
			\mathbf{0}  & \mathbf{I}  &\mathbf{F}_v^R\\
			\mathbf{0}  & {\mathbf{0}}  &\mathbf{F}_R^R\\
		\end{bmatrix}
		\mathbf{F}_\mathbf{B} &= \begin{bmatrix}
			\mathbf{0}  & \mathbf{0} & \mathbf{0}  & \mathbf{0}  & \mathbf{0}\\
			\mathbf{F}_{v}^{b_a}  & \mathbf{0} & \mathbf{0}  & \mathbf{0}  & {\mathbf{I}\mathrm{\Delta}t}\\
			\mathbf{0}  & \mathbf{F}_{R}^{b_\omega} & \mathbf{0}  & \mathbf{0}  & \mathbf{0}\\
		\end{bmatrix}
\end{align}
where $\mathbf{F}_v^R = - {{}^{G}\mathbf{R}_{I_{i}}^{~}}\left( {\mathbf{a}}_{m} \right)_{\times}\mathrm{\Delta}t$, 
$\mathbf{F}_{v}^{b_a}  = - {{}^{G}\mathbf{R}_{I_{i}}^{~}}\mathrm{\Delta}t$,
$\mathbf{F}_{R}^{R}  = exp\left( {- {\boldsymbol{\omega}}_{m}\Delta t} \right)$, 
$\mathbf{I} = \mathbf{I}_{3\times3}$, 
$\mathbf{F}_{R}^{b_\omega}  = -\mathbf{A}\left( {\boldsymbol{\omega}}_{m}{\Delta}t \right)^T{\Delta}t$ in which $\mathbf{A}\left(\cdot \right)^{-1}$ is defined in \cite{xu2022fast}.

\subsection{Ego-Velocity Measurements}
The ego-velocity ${}^{R}\mathbf{v}_{m}^{~}$ in the radar frame estimated from a radar scan can serve as the velocity measurement,
\begin{align}
	\mathbf{r}_{v} &= {{}^{R}\mathbf{v}^{~}} - {{}^{R}\mathbf{v}_{m}^{~}} \approx {\widehat{\mathbf{r}}}_{v} + \mathbf{H}_{v}\widetilde{\mathbf{x}} \\
		{{}^{R}\mathbf{v}^{~}} &= {{}^{I}\mathbf{R}_{R}^{T}}\left( {{}^{G}\mathbf{R}_{I}^{T}}{{}^{G}\mathbf{v}_{I}^{~}} + \left( {\boldsymbol{\omega}_{m} - \mathbf{b}_{\omega}^{~}} \right)_{\times} {{}^{I}\mathbf{l}_{R}^{~}} \right)
\end{align}
where ${\widehat{\mathbf{r}}}_{v}$ is the residual and $\mathbf{H}_{v}$ is the Jacobian matrix of ${\widehat{\mathbf{r}}}_{v}$ relative to the error state, as derived in \cite{doer2021yaw}.

%\begin{equation}
%	\begin{aligned}
%		\mathbf{H}_{v} = 
%		\begin{bmatrix}
%			\mathbf{0}_{3}  &\mathbf{H}_{v}^{v} & \mathbf{H}_{v}^{R} & \mathbf{0}_{3} & \mathbf{H}_{v}^{{\mathbf{b}}_{\omega}} & \mathbf{H}_{v}^{R_e} & \mathbf{H}_{v}^{l} & \mathbf{0}_{3} \\
%		\end{bmatrix}
%	\end{aligned}
%\end{equation}
%where $\mathbf{v}_{\omega} = \left( {\boldsymbol{\omega}_{m} - {\widehat{\mathbf{b}}}_{\omega}^{~}} \right)_{\times}{{}^{I}{\widehat{\mathbf{l}}}_{R}^{~}}$,
%$\mathbf{H}_{v}^{v} = {{}^{I}{\widehat{\mathbf{R}}}_{R}^{T}}{{}^{G}{\widehat{\mathbf{R}}}_{I}^{T}}$,

%$\mathbf{H}_{v}^{R} = - {{}^{I}{\widehat{\mathbf{R}}}_{R}^{T}}\left( {{{}^{G}{\widehat{\mathbf{R}}}_{I}^{T}}{{}^{G}{\widehat{\mathbf{v}}}_{I}^{~}}} \right)_{\times}$,
%
%$\mathbf{H}_{v}^{R_e} = - \left( {{{}^{I}{\widehat{\mathbf{R}}}_{R}^{T}}\left( {{{}^{G}{\widehat{\mathbf{R}}}_{I}^{T}}{{}^{G}{\widehat{\mathbf{v}}}_{I}^{~}} + \mathbf{v}_{\omega}} \right)} \right)_{\times}$,
%
%$\mathbf{H}_{v}^{{\mathbf{b}}_{\omega}} = - {{}^{I}{{\widehat{\mathbf{R}}}_{R}^{T}\left( {{}^{I}{\widehat{\mathbf{l}}}_{R}^{~}} \right)_{\times}}}$,
%$\mathbf{H}_{v}^{l} = - {{}^{I}{\widehat{\mathbf{R}}}_{R}^{T}}\left( {\boldsymbol{\omega}_{m} - {\widehat{\mathbf{b}}}_{\omega}^{~}} \right)_{\times}$.

We prune those 3D ego-velocity measurements with the Mahalanobis distance of $\mathbf{r}_v$ $>\chi_{3, 0.05}^2$ value at significance level 0.05.
In the update step, the Kalman gain is calculated as
\begin{equation}
	\mathbf{K}_{v} = \mathbf{P}{\mathbf{H}_{v}}^{T}\left( {\mathbf{H}_{v}\mathbf{P}{\mathbf{H}_{v}}^{T} + \mathbf{R}_{v}} \right)^{- 1}
\end{equation}
where $\mathbf{P}$ is the covariance of the state $\mathbf{x}$ and $\mathbf{R}_{v}$ is the measurement covariance matrix.
The updated state ${\bar{\mathbf{x}}}_{v}$ and covariance ${\bar{\mathbf{P}}}_{v}$ are
\begin{align}
		{\bar{\mathbf{x}}}_{v} & = \widehat{\mathbf{x}} \oplus \widetilde{\mathbf{x}}  = \widehat{\mathbf{x}} \oplus (-\mathbf{K}_{v}{\widehat{\mathbf{r}}}_{v}) \\
	{\bar{\mathbf{P}}}_{v} &= \left( \mathbf{I} - \mathbf{K}_{v}\mathbf{H}_{v} \right)\mathbf{P}
\end{align}
where $\oplus$ denotes the state plus operation, in which  
${^{G}\widehat{\mathbf{R}}_{I_{i + 1}}} \oplus {^{G}\widetilde{\mathbf{R}}_{I_{i + 1}}} = {^{G}\widehat{\mathbf{R}}_{I_{i + 1}}}exp({^{G}\widetilde{\mathbf{R}}_{I_{i + 1}}})$,
${^{I_{i + 1}}\widehat{\mathbf{R}}_{R_{i + 1}}} \oplus {^{I_{i + 1}}\widetilde{\mathbf{R}}_{R_{i + 1}}}= {^{I_{i + 1}}\widehat{\mathbf{R}}_{R_{i + 1}}}exp({^{I_{i + 1}}\widetilde{\mathbf{R}}_{R_{i + 1}}})$, 
and the others follow the common plus operation.

\subsection{Scan-to-Submap Matching Constraints}
We also use point cloud matching to achieve higher accuracy for positioning and mapping.
Odometry methods based on radar scans usually adopt two types of matching: scan-to-scan or scan-to-submap. 
Scan-to-scan matching derives point matches from two sequential scans,
thus, involves fewer points and relatively lower computation.
Otherwise, scan-to-scan matching is less robust and prone to error accumulation with sparse point clouds or substantial changes in local scenes (turning, up-slope, etc.).
Scan-to-submap matching maintains a local map to match the current scan.
According to the maintenance methods, 
typical local maps include key-frame sliding window maps, e.g., KSWF \cite{huai2022observability}, and 
incrementally updated grid maps, e.g., ikd-tree \cite{xu2022fast}.
Compared with scan-to-scan matching, scan-to-submap matching has higher stability and accuracy.
Thus, we adopt scan-to-submap matching based on the ikd-tree which facilitates real-time computation.

The spatial point matches from scan-to-submap matching are used to update the state in an iterative Kalman filter.
The distance of a point match is usually measured by point-to-point, point-to-distribution, or distribution-to-distribution distances\cite{koide2021voxelized}.
The distribution-to-distribution metric considers the local geometry of the source and the target point to weight the distance residuals.
Thus, it is resilient to data noise and suitable for 
the sparse radar scans.

Due to the sparsity of 4D radar scans, it is difficult to ensure that the points in the current scan and the local map have exact one-to-one matches.
Thus, we consider the $N$ nearest neighbor points $^{G}\mathbf{b}_{j}$ ($j=1, \ldots, N, N=5$ in our tests) 
in the submap of a point ${^{R}\mathbf{a}}$ in the current scan, 
weight the distance between $\{^{G}\mathbf{b}_{j}\}$'s centroid and ${^{R}\mathbf{a}}$ by their covariance to achieve a distribution-to-multi-distribution effect.
The weighting scheme empirically improves the registration of a scan to a submap.
The weighted point match distance is computed as
\begin{align}
	\mathbf{r}_p &= \mathbf{G}_p ( {\Sigma_j({{}^{G}\mathbf{b}_{j}}}/N) - {{}^{G}\mathbf{T}_{R}^{~}} \cdot {{}_{\mathbf{~}}^{R}\mathbf{a}} ) \approx {\widehat{\mathbf{r}}}_p + \mathbf{H}_p\widetilde{\mathbf{x}} \\
	{{}^{G}\mathbf{T}_{R}^{~}} &= {{}^{G}\mathbf{T}_{I}^{~}}{{}^{I}\mathbf{T}_{R}^{~}} = 
	\begin{bmatrix}
		{{}^{G}\mathbf{R}_{I}^{~}} & {{}^{G}\mathbf{l}_{I}^{~}} \\
		\mathbf{0}_{1 \times 3} & 1 \\
	\end{bmatrix}
	\begin{bmatrix}
		{{}^{I}\mathbf{R}_{R}^{~}} & {{}^{I}\mathbf{l}_{R}^{~}} \\
		\mathbf{0}_{1 \times 3} & 1 \\
	\end{bmatrix} \\
	\mathbf{G}_p &= 
	\sqrt{({\Sigma_{j} (\mathbf{C}_{j}^{B}} / N) + {^{G}{\widehat{\mathbf{T}}}_{R}} \cdot \mathbf{C}^{A} \cdot {^{G}{\widehat{\mathbf{T}}}_{R}^{T}})^{-1}} \\
	{\widehat{\mathbf{r}}}_p &= \mathbf{G}_p ( {{\Sigma_j ({^{G}\mathbf{b}_{j}}}/N) - {^{G}{\widehat{\mathbf{T}}}_{R}^{~}} \cdot {{}_{\mathbf{~}}^{R}\mathbf{a}}} ) \\
	\mathbf{H}_p &=
	\begin{bmatrix}
		\mathbf{G}_p & \mathbf{0}_{3} & \mathbf{H}_p^{R} & \mathbf{0}_{3} & \mathbf{0}_{3} & \mathbf{H}_p^{R_e} & \mathbf{H}_p^{l} & \mathbf{0}_{3} \\
	\end{bmatrix}
\end{align}
where $\mathbf{C}^{A}$ and $\mathbf{C}^{B}_{j}$ are the spatial covariance of the five points close to ${{}_{\mathbf{~}}^{R}\mathbf{a}}$ and $^{G}\mathbf{b}_{j}$ (including themselves), resp., 
and $\mathbf{G}_p$ is the weight matrix. ${{}^{G}\mathbf{T}_{R}^{~}}$ is the transformation matrix from radar coordinate system to global coordinate system.
${\widehat{\mathbf{r}}}_p$ and $\mathbf{H}_p$ are the computed residual and the Jacobian matrix relative to the state, in which $\mathbf{H}_p^{R} = \mathbf{G}_p\left( {{{}^{G}\mathbf{R}_{I}^{~}}\left( {{{}^{I}\mathbf{R}_{R}^{~}} \cdot {{}_{\mathbf{~}}^{R}\mathbf{a}} + {{}^{I}\mathbf{l}_{R}^{~}}} \right)} \right)_{\times}$, 
$\mathbf{H}_p^{R_e} = \mathbf{G}_p{{}^{G}\mathbf{R}_{I}^{~}}\left( {{{}^{I}\mathbf{R}_{R}^{~}} \cdot {{}_{\mathbf{~}}^{R}\mathbf{a}}} \right)_{\times}$, $\mathbf{H}_p^{l} =- \mathbf{G}_p{{}^{G}\mathbf{R}_{I}^{~}}$.

During the iterative Kalman filtering update, 
$\mathbf{x}$ should be kept nearly to the initial estimate $\widehat{\mathbf{x}}$, namely
\begin{equation}
	\mathbf{x} \ominus \widehat{\mathbf{x}} = \left( {{\widehat{\mathbf{x}}}^{\kappa} \oplus {\widetilde{\mathbf{x}}}^{\kappa}} \right) \ominus \widehat{\mathbf{x}} = 
	\left( {\widehat{\mathbf{x}}}^{\kappa} \ominus \widehat{\mathbf{x}} \right) + \mathbf{J}^{\kappa}{\widetilde{\mathbf{x}}}^{\kappa} \approx \mathbf{0}
\end{equation}
\begin{equation}
	\begin{aligned}
		\mathbf{J}^{\kappa} = \begin{bmatrix}
			\mathbf{I}_{6} & \mathbf{0}_{6 \times 3} & \mathbf{0}_{6} & \mathbf{0}_{6 \times 3} & \mathbf{0}_{6} \\
			\mathbf{0}_{3 \times 6} & \mathbf{J}_{1}^{\kappa} & \mathbf{0}_{3 \times 6} & \mathbf{0}_{3} & \mathbf{0}_{3 \times 6} \\
			\mathbf{0}_{6} & \mathbf{0}_{6 \times 3} & \mathbf{I}_{6} & \mathbf{0}_{6 \times 3} & \mathbf{0}_{6} \\
			\mathbf{0}_{3 \times 6} & \mathbf{0}_{3} & \mathbf{0}_{3 \times 6} & \mathbf{J}_{2}^{\kappa} & 0_{3 \times 6} \\
			\mathbf{0}_{6} & \mathbf{0}_{6 \times 3} & \mathbf{0}_{6} & \mathbf{0}_{6 \times 3} & \mathbf{I}_{6} \\
		\end{bmatrix}
	\end{aligned}
\end{equation}
where ${\widehat{\mathbf{x}}}^{\kappa}$ is the $\kappa$th filter update value,
${\widetilde{\mathbf{x}}}^{\kappa}$ is the corresponding error state, 
and $\mathbf{J}^{\kappa}$ is the linearized Jacobi matrix (at ${\widetilde{\mathbf{x}}}^{\kappa} = \mathbf{0}$), 
$\mathbf{J}_{1}^{\kappa} = \mathbf{~}\mathbf{A}\left( {{}^{G}{\widehat{\mathbf{R}}}_{I}^{\kappa}} \ominus {{}^{G}{\widehat{\mathbf{R}}}_{I}^{~}} \right)^{- T}$, 
$\mathbf{J}_{2}^{\kappa} = \mathbf{~}{\mathbf{A}\left( {{}^{I}{\widehat{\mathbf{R}}}_{R}^{\kappa}} \ominus {{}^{I}{\widehat{\mathbf{R}}}_{R}^{~}} \right)}^{- T}$.

Then the update process can be described as a MAP estimation problem as follows
\begin{equation}
	\begin{aligned}
		\min\limits_{{\widetilde{\mathbf{x}}}^{\kappa}}
		\left\| {\left( {\widehat{\mathbf{x}}}^{\kappa} \ominus \widehat{\mathbf{x}} \right)  +  \mathbf{J}^{\kappa}{\widetilde{\mathbf{x}}}^{\kappa}} \right\|_{{\widehat{\mathbf{P}}}^{- 1}}^{2}
		+ {\sum_{j = 1}^{m}\left\| {{{}^{j}{\widehat{\mathbf{r}}}_p^{\kappa}} +
				{{}^{j}\mathbf{H}_p^{\kappa}}{\widetilde{\mathbf{x}}}^{\kappa}} \right\|_{{}_{\mathbf{~}}^{j}\mathbf{R}_p^{- 1}}^{2}}
	\end{aligned}
\end{equation}
where $\widehat{\mathbf{x}} = {\bar{\mathbf{x}}}_{vel}$, 
$\widehat{\mathbf{P}} = {\bar{\mathbf{P}}}_{vel}$ is the state and covariance matrix after ego-velocity update, 
and ${}_{\mathbf{~}}^{j}\mathbf{R}_p^{- 1}$ is the noise covariance for the $j$-th point.

Solving the above problem, we obtain
\begin{equation}
	\mathbf{K}_p^{\kappa} = \mathbf{P}^{\kappa}\left( \mathbf{H}_p^{\kappa} \right)^{T}\left( { \mathbf{H}_p^{\kappa} \mathbf{P}^{\kappa}\left( \mathbf{H}_p^{\kappa} \right)^{T} +  \mathbf{R}_p } \right)^{-1}
\end{equation}
\begin{equation}
		{\widehat{\mathbf{x}}}^{\kappa + 1} =  {\widehat{\mathbf{x}}}^{\kappa}  \oplus
		[- \mathbf{K}_p^{\kappa}{\widehat{\mathbf{r}}}_p^{\kappa} - \left( \mathbf{I} - 
		\mathbf{K}_p^{\kappa}\mathbf{H}_p^{\kappa} \right){\mathbf{J}^{\kappa}}^{-1}\left( {\widehat{\mathbf{x}}}^{\kappa} \ominus \widehat{\mathbf{x}} \right)]
\end{equation}
where $\mathbf{H}_p^{\kappa}= \left\lbrack {{}^{1}\mathbf{H}_p^{\kappa}},{{}^{2}\mathbf{H}_p^{\kappa}},\ldots,{{}^{m}\mathbf{H}_p^{\kappa}} \right\rbrack$, 
$\mathbf{R}_p= diag\left( {{}_{\mathbf{~}}^{1}\mathbf{R}_p},{{}_{\mathbf{~}}^{2}\mathbf{R}_p},\ldots,{{}_{\mathbf{~}}^{m}\mathbf{R}_p} \right)$, 
$\mathbf{P}^{\kappa} = \left( \mathbf{J}^{\kappa} \right)^{- 1}\widehat{\mathbf{P}}\left( \mathbf{J}^{\kappa} \right)^{- T}$.

When the iterations converge or reach the maximum iterations,
the projection residual optimization state ${\bar{\mathbf{x}}}_p$ and covariance ${\bar{\mathbf{P}}}_p$ can be obtained as
\begin{align}
	{\bar{\mathbf{x}}}_p &= {\widehat{\mathbf{x}}}^{\kappa + 1} \\
	{\bar{\mathbf{P}}}_p &= \left( \mathbf{I} - \mathbf{K}_p^{\kappa} \mathbf{H}_p^{\kappa} \right)\mathbf{P}^{\kappa}
\end{align}
where ${\widehat{\mathbf{x}}}^{\kappa + 1}$ is the filter state of the $\kappa + 1$ iteration and
$\mathbf{K}_p^{\kappa}$ is the Kalman gain calculated from the $\kappa$ iteration.

\section{Experiments}
\subsection{In-house Data}
\begin{figure}[thpb]
	\centering
	\includegraphics[width=5.2cm]{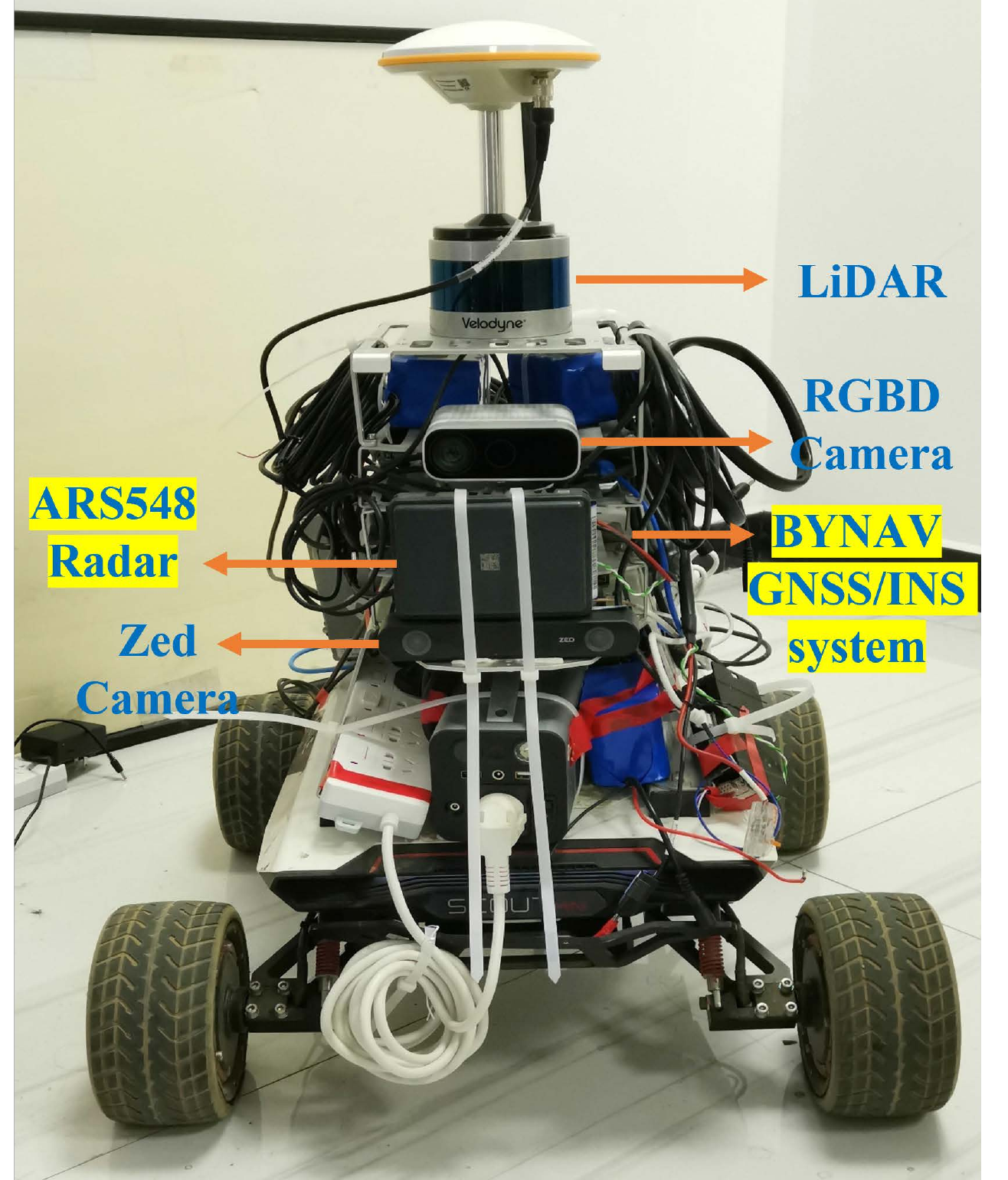}
	\caption{The ground robot and the sensors for tests. Note that the ZED camera and the RGBD camera are not used.}
	\label{fig:setup}
\end{figure}

We equipped a ground robot with a 4D radar ARS548 from Continental, 
which scans at 15 Hz in the $76-77$ GHz, with an elevation angle of view (AOV) $\pm 20{^\circ}$ and 
an azimuth AOV $\pm 60{^\circ}$, azimuth angle resolution $0.2{^\circ}$, elevation angle resolution $0.1{^\circ}$, detection range $\sim$300 m, and distance accuracy 0.3 m.
The reference trajectory is obtained by the Bynav X1-5H GNSS/INS module in open sky areas.
The data from the Bynav module's built-in IMU (EPSON G345) are fused with the radar data.
The IMU has a gyroscope bias stability $0.00075$ ${^\circ}/s$, angular random walk $0.003$ ${^\circ}/\sqrt{s}$, accelerometer bias stability $70$ $\mu g$, 
and the velocity random walk $0.0005$ $m/\sqrt{s^3}$.
The robot and sensors are shown in Fig.~\ref{fig:setup}.

\begin{table}[thpb]
	\centering
	\caption{Statistics of EKFRIO, iRIO, and iRIOM on our sequence 1-3}
	\label{tab:addlabel1}%
	\begin{threeparttable}
		\begin{tabular}{cccccccc}
			\toprule
			\multicolumn{1}{c}{\multirow{2}[3]{*}{}}  & \multirow{2}[3]{*}{Method}  & \multicolumn{2}{c}{Closure Error}  & \multicolumn{2}{c}{APE RMSE} & \multicolumn{2}{c}{RPE RMSE} \\
			\cmidrule(r){3-4}      \cmidrule(r){5-6}   \cmidrule(r){7-8}     
			&       & Hor       & Ver            & Trans       & Rot        & Trans      & Rot\\
			\cmidrule{1-8} 
			\multicolumn{1}{c}{\multirow{3}[0]{*}{1}} 
			& EKFRIO           & 0.530   & 15.759      & 5.616  & 10.261       & 0.055   & 0.183\\
			& iRIO             & 0.029   & 0.010       & 0.313  & 2.923        & 0.040   & 0.171\\
			& iRIOM            & 0.027   & 0.010       & 0.305  & 2.751        & 0.040   & 0.171\\
			& FastLIO     & 0.024   & 0.007       & 0.147  & 5.331    & 0.025  & 0.343 \\
			\cmidrule{1-8} 
			\multicolumn{1}{c}{\multirow{3}[0]{*}{2}} 
			& EKFRIO           & 17.194    & 24.631      & 10.068  & 9.859     & 0.026   & 0.120\\
			& iRIO             & 1.517     & 5.073       & 1.518  & 2.866     & 0.022   & 0.114\\
			& iRIOM            & 0.279     & 0.001       & 0.336  & 2.626    & 0.022   & 0.112\\
			& FastLIO     & 0.151     & 0.025       & 0.372  & 6.363   & 0.020  & 0.322\\
			\cmidrule{1-8} 
			\multicolumn{1}{c}{\multirow{3}[0]{*}{3}} 
			& EKFRIO          & 3.741     & 9.268       & 3.388  & 8.655     & 0.183   & 0.458\\
			& iRIO            & 0.051     & 0.032       & 0.303  & 8.368     & 0.132   & 0.343\\
			& iRIOM           & 0.050     & 0.032       & 0.307  & 8.359     & 0.131  & 0.340\\
			& FastLIO    & 0.012     & 0.007       & --  & --    & --  & -- \\
			\bottomrule
		\end{tabular}%
		\begin{tablenotes}
			\item[a] APE Rot. RMSEs are in $^\circ$, APE Trans. RMSEs and Closure Errors are in m. RPE Rot. RMSEs are in $^\circ/m$, and RPE Trans. RMSEs are in $\%$.
			\item[b] FastLIO denotes FastLIO-SLAM with loop closure. GNSS/INS results was unreliable for sequence 3, so we took the FASTLIO-SLAM trajectory as reference; thus, the RMSEs of FastLIO-SLAM are zero.
		\end{tablenotes}
	\end{threeparttable}
\end{table}%

We collected data in two outdoor scenes and one indoor scene.
We began the data collection at an area with open sky and performed maneuvers of straight lines and then figure eight curves to make sure the GNSS RTK and the INS system is well aligned, 
so that we have an accurate reference.
The reported accuracy of the Bynav module is $\sim$3 cm with profuse RTK data, but drops to 25 cm with a 10-second RTK outage.

Sequence 1 is at an outdoor basketball court surrounded by trees.
The 6-minute data has a maximum velocity of 3 m/s and large angular velocity at turns.
The whole session has the centimeter-level GNSS RTK/INS solution as the reference, and we also started and stopped at the same pose.

Sequence 2 is collected by traversing around the Xinghu Building in Wuhan University for two laps.
There are numerous vehicles, pedestrians, and other moving targets. 
The 9-minute data has a maximum speed about 2 m/s and angular speed at turns is small. 
Despite occasional short GNSS blockages due to high-rises, the INS poses were always reported as good, and the trajectory started and stopped at the same pose.

Sequence 3 is an underground parking lot with narrow passageways and serious occlusions. 
The ground and walls in the scene are very smooth, leading to multipath reflections and a large number of ``ghost" points. 
The 5-minute data has a maximum speed of 2 m/s and small angular speed at turns. 
The GNSS/INS solution was somewhat unreliable, but the trajectory started and stopped at the same pose.

\subsection{In-house Data Results}
For all three sequences, we evaluate the algorithm performance by the trajectory closure error (CE).
To mitigate randomness, we run an algorithm 10 times on each sequence and calculated the mean value of the closure error.
Moreover, we evaluate the accuracy of the methods by calculating the difference between the estimated trajectory and the reference.
For sequence 1 and 2, we take the centimeter-level trajectory obtained by the fusion of GNSS and IMU as reference, 
and evaluate the pose estimation accuracy by calculating the root mean square error (RMSE) between the reference trajectory and the algorithm results,
include the absolute pose error (APE) and the relative pose error (RPE).
For sequence 3, we obtain a rough reference trajectory by fusing lidar and IMU data by FastLIO-SLAM \cite{xu2022fast}.

We conduct comparison with the EKFRIO method in \cite{doer2020ekf} and a recent lidar-inertial SLAM method with loop closure, FastLIO-SLAM \cite{xu2022fast}. 
EKFRIO is the best-performing open-source 4D radar inertial odometry method known to us.
Lidar data for FastLIO-SLAM was obtained from the Velodyne VLP-16 on the robot.
We tested two variants of iRIOM: the iRIO without loop closure and the iRIOM. 
The accuracy statistics on our three sequences are shown in Table~\ref{tab:addlabel1}.
We see that the position accuracy of iRIO is much better than that of EKFRIO, 
and iRIOM further reduces the drift in the Z-direction by loop closure.
Also, the accuracy of iRIOM is comparable to that of FastLIO-SLAM on these sequences.

\begin{figure}[thpb]
	\centering
	\subfloat{
		\includegraphics[width=0.5\linewidth]{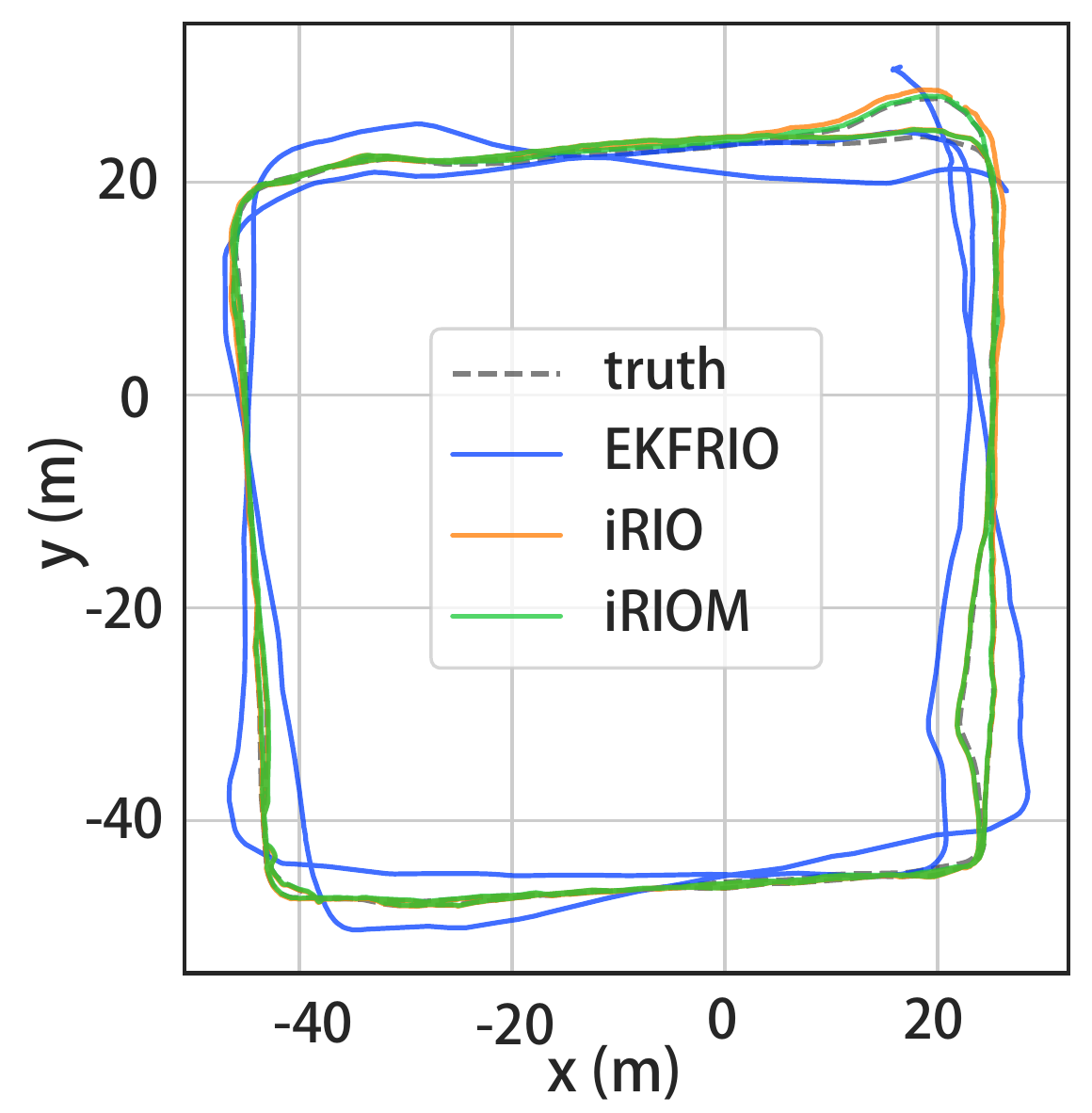}} \vspace{-0cm}
	\subfloat{
		\includegraphics[width=1\linewidth]{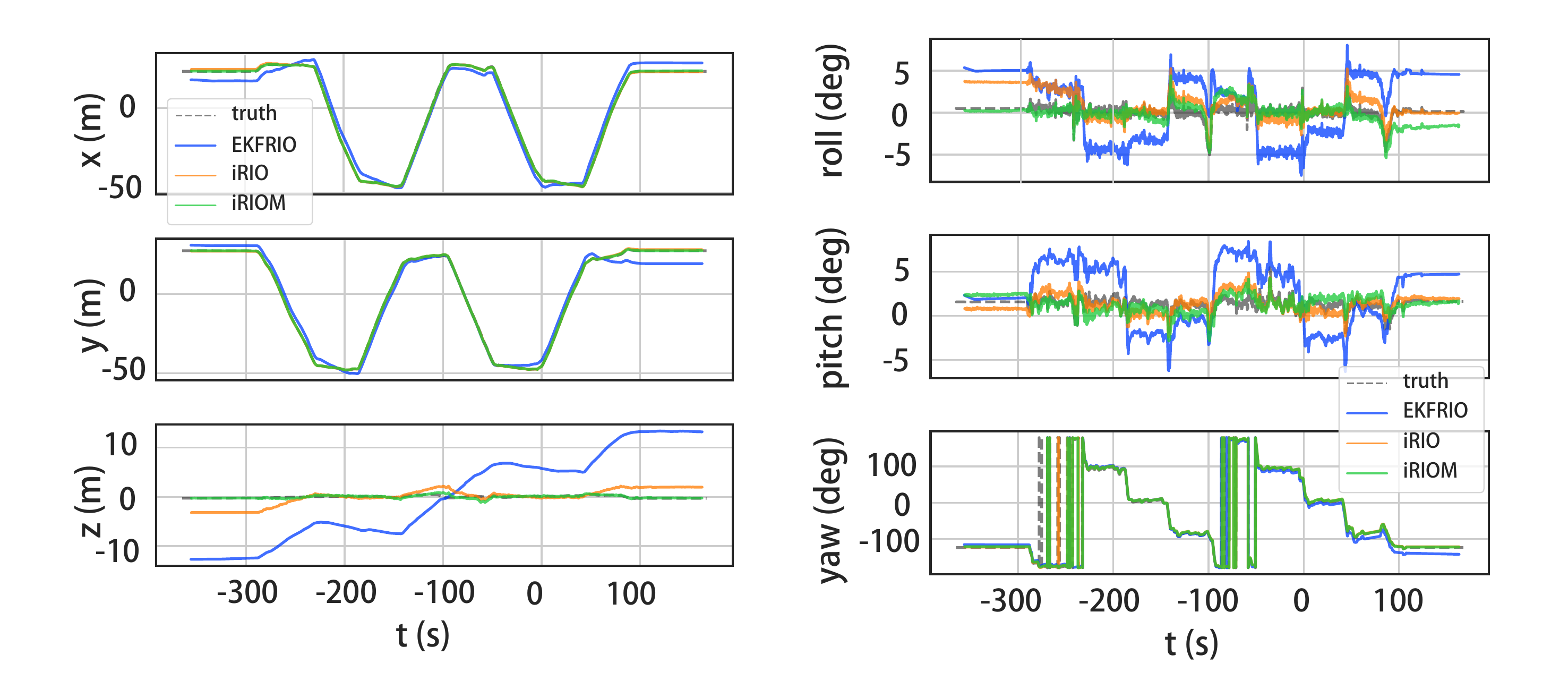}}
	\caption{Top: Trajectories estimated by the EKFRIO, iRIO and iRIOM on our sequence 2.
		Bottom: The position and attitude of estimated trajectories by these methods aligned to the reference.}
	\label{fig:assign}
\end{figure}

\begin{figure}[thpb]
	\centering
	\includegraphics[width=8.5cm]{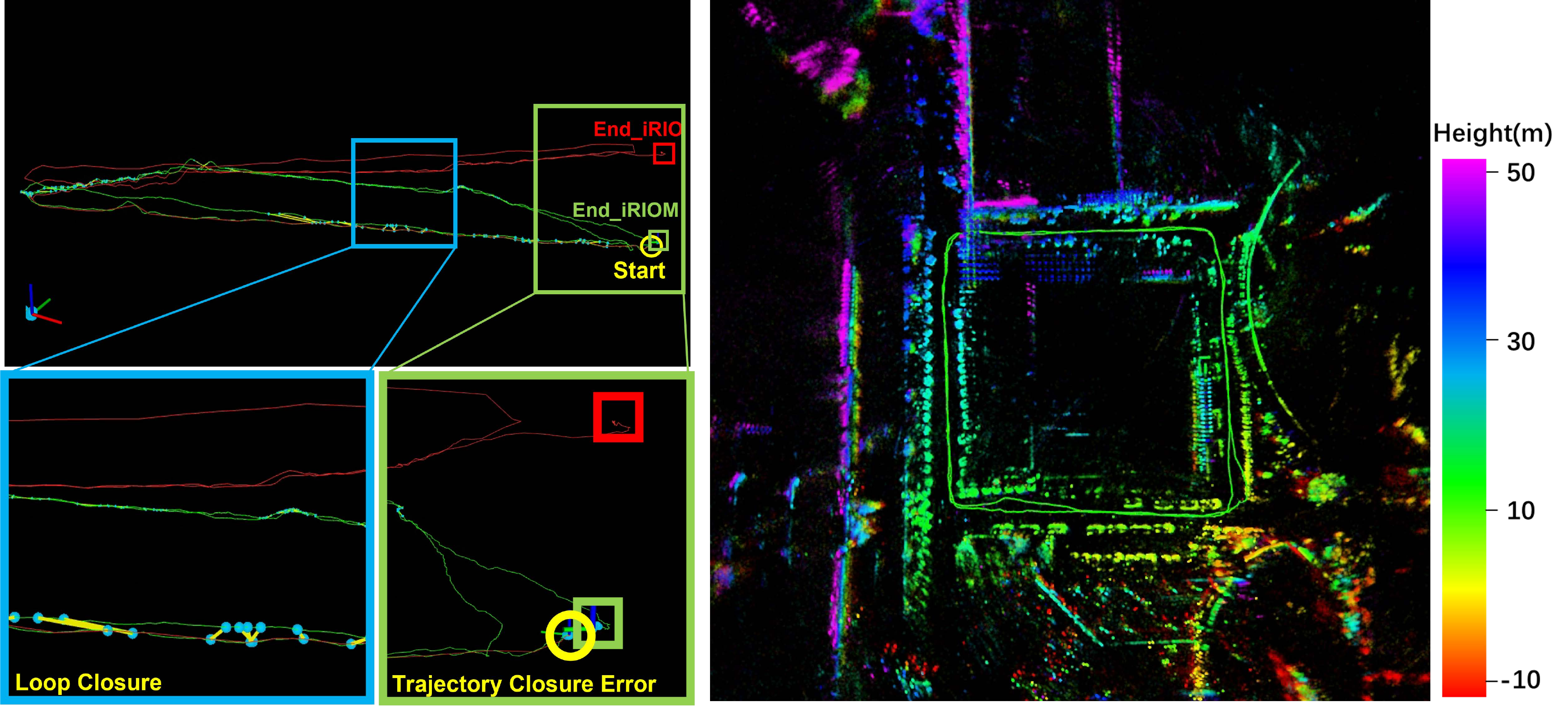}
	\caption{The detected loop closure and closure error (left), 
	and the mapping result of iRIOM (right) on our sequence 2 where points are colored by height.}
	\label{fig:closure error}
\end{figure}

For sequence 2, Fig.~\ref{fig:assign} shows the trajectories estimated by the three radar methods along with the ground truth.
Fig.~\ref{fig:closure error} shows the trajectory closure error of iRIO on sequence 2.
Both figures show that by using scan-to-submap point matching, the trajectory closure error and trajectory RMSEs of iRIO are significantly lower than those of EKFRIO.
However, due to the large scene of sequence 2, iRIO still has a large drift in the Z-direction. 
The iRIOM method further refines the position and attitude results by loop closure, thus, 
the trajectory closure error is reduced by an order of magnitude.

\begin{figure}[thpb]
	\centering
	\subfloat{
		\includegraphics[width=0.55\linewidth]{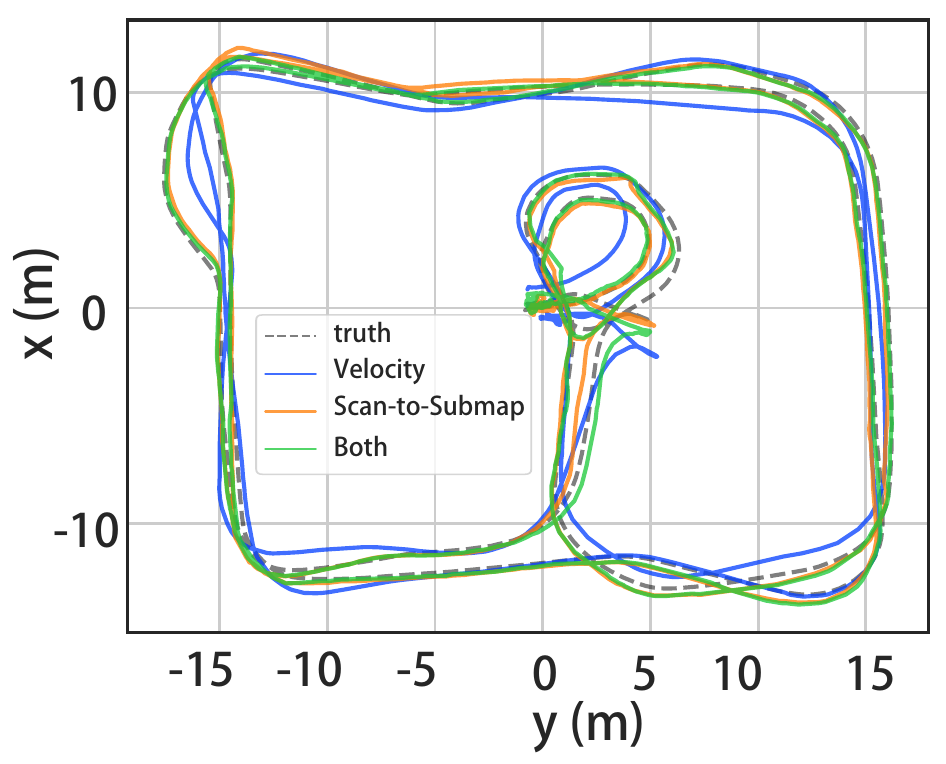}} \vspace{-0.0cm}
	\subfloat{
		\includegraphics[width=1\linewidth]{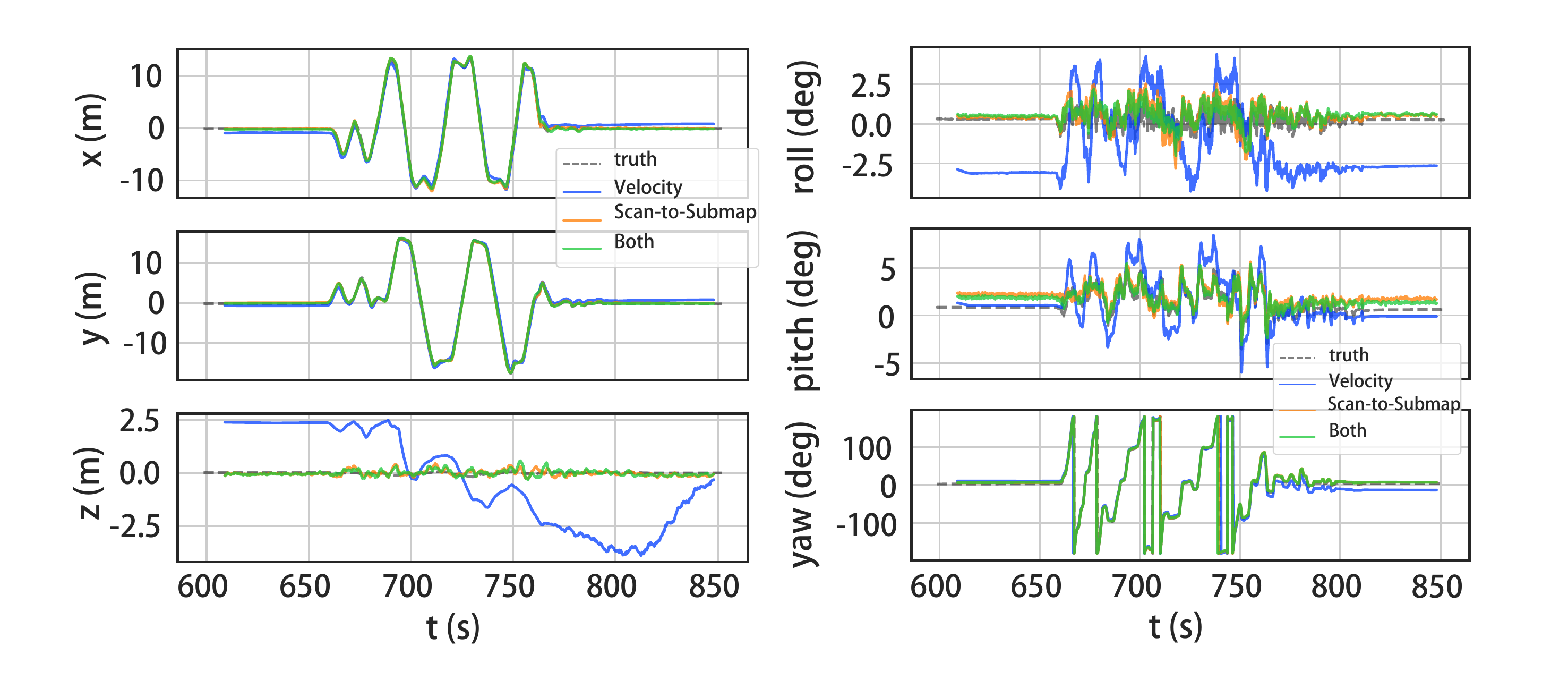}}
	\caption{Top: Trajectories of variants of iRIOM described in Section~\ref{subsec:ablation} on our sequence 1. 
	Velocity: Only using ego-velocity for filter update; Scan-to-submap: only using scan-to-submap matches for filiter update; Both: both measurements are used for update.
	Bottom: The position and attitude of estimated trajectories by these variants aligned to the reference.}
	\label{fig:assign2}
\end{figure}

\begin{table}[!thpb]
	\centering
	\caption{Timing statistics(max/min and mean in ms) of iRIOM on our sequence 1-3}
	\label{tab:addlabel3}%
	\begin{threeparttable}
		\begin{tabular}{ccccc}
			\toprule
			\cmidrule{1-5} 
			& IMU predict   & \makecell[c]{Ego vel. \\ update}      & \makecell[c]{Scan-to-Submap \\ update}  & Total    \\ 
			\cmidrule{1-5} 
			1             & \makecell[c]{10.66 / 0.09\\0.24}   & \makecell[c]{13.42 / 0.05\\0.47}       & \makecell[c]{379.18 / 1.47\\48.69}  & \makecell[c]{380.08 / 1.72\\49.41}     \\
			\cmidrule{1-5} 
			2            & \makecell[c]{9.53 / 0.08\\0.19}   & \makecell[c]{6.24 / 0.06\\0.51}       & \makecell[c]{284.55 / 2.28\\50.07}  & \makecell[c]{285.10 / 3.13\\50.77}     \\
			\cmidrule{1-5} 
			3            & \makecell[c]{4.89 / 0.08\\0.18}   & \makecell[c]{2.84 / 0.05\\0.36}       & \makecell[c]{119.29 / 0.34\\12.87}  & \makecell[c]{120.55 / 0.75\\13.41}     \\
			\bottomrule
		\end{tabular}%
%		\begin{tablenotes}
%			\item[a] The table shows the Max/Min/Mean time consumption to process a single radar scan, in milliseconds.
%		\end{tablenotes}
	\end{threeparttable}
\end{table}%

We also calculated the running time of our algorithm shown in Table~\ref{tab:addlabel3}.
The iRIOM method can run in real time on a consumer laptop processing our ARS548 dataset with $\sim15$ Hz radar scans.
The most costly part of iRIOM is the scan-to-submap matching which includes the iterative EKF update.
The running times are also related to the scene as iRIOM runs much faster indoors than outdoors.
We believe this is caused by the submap maintenance.

\subsection{Ablation Studies}
\label{subsec:ablation}
To verify the effect of the IMU data, ego-velocity constraints, scan-to-submap matching constraints,
we tested ablated variants of iRIOM on our three sequences.
The benefit of the IMU data is validated against a constant velocity model for prediction. 
We also compare the scan-to-submap matching with a temporal scan-to-scan matching. 
Note that we disable the loop closure module to highlight the effect of each type of constraints.

\begin{figure}[!thpb]
	\centering
	\includegraphics[width=0.95\columnwidth]{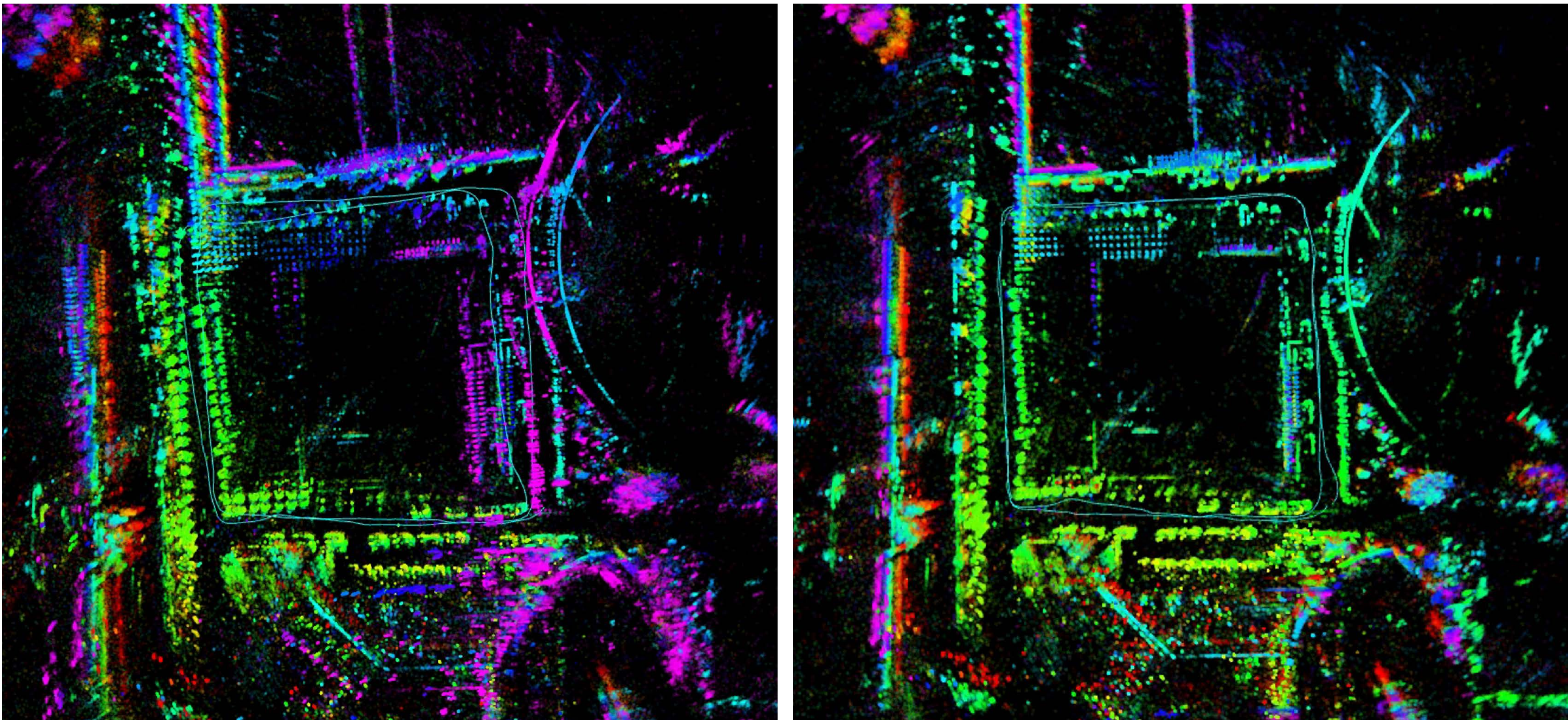}
	\caption{The mapping results of iRIO with scan-to-five-scans matching and with the default scan-to-submap matching.}
	\label{fig:scan-to-scans}
\end{figure}

For variants of iRIOM,
Table~\ref{tab:addlabel2} lists the trajectory closure errors and trajectory RMSEs.
Fig.~\ref{fig:assign2} shows the trajectory estimation results of the ablated iRIOM with either the ego-velocity or the scan-to-submap matches on our sequence 1.
Also shown in Fig.~\ref{fig:mapping-data1}, quantitatively, the mapping results with only ego-velocity constraints still have apparent blur and vertical drift. 
The point matches helped better estimate the map points and significantly reduce the mapping blur.
Fig.~\ref{fig:scan-to-scans} shows the mapping results of the iRIO variant with scan-to-five-scans matching.

\begin{table}[thpb]
	\centering
	\caption{Ablation Test Statistics on Our Sequence 1-3}
	\label{tab:addlabel2}%
	\begin{threeparttable}
		\begin{tabular}{ccccccccc}
			\toprule
			\multicolumn{1}{c}{\multirow{2}[3]{*}{}} & \multirow{2}[3]{*}{Meas.}      & \multirow{2}[3]{*}{F} & \multicolumn{2}{c}{Closure Error} & \multicolumn{2}{c}{RMSE APE}  & \multicolumn{2}{c}{RMSE RPE}\\
			\cmidrule(r){4-5}      \cmidrule(r){6-7}   \cmidrule(r){8-9}
			&       &       & Hor      & Ver             & Trans       & Rot       & Trans      & Rot\\
			\cmidrule{1-9} 
			\multicolumn{1}{c}{\multirow{5}[0]{*}{1}} 
			& Vel.                  & 1  & 2.398 & 2.606     & 2.327 & 9.744     & 0.046 & 0.229 \\
			& S2M                   & 4  & 0.023 & 0.091     & 0.342 & 2.926      & 0.042 & 0.172  \\
			& Both                  & 0 & 0.029 & 0.010     & 0.313  & 2.923      & 0.040 & 0.171  \\
			& CVM                   & 5  & 25.598 & 1.433     & 9.106 & 8.591     & 0.121 & 0.300 \\
			& S2S                   & 2  & 3.402 & 1.314     & 1.441 & 5.398      & 0.043 & 0.277  \\
			\cmidrule{1-9} 
			\multicolumn{1}{c}{\multirow{5}[0]{*}{2}} 
			& Vel.                  & 1  & 25.422 & 8.913    & 3.367 & 13.298      & 0.022 & 0.172 \\
			& S2M                   & 1  & 1.540  & 5.417    & 1.556  & 2.981      & 0.023 & 0.172 \\
			& Both                  & 0 & 1.517  & 5.073    & 1.518  & 2.866      & 0.022 & 0.114 \\
			& CVM                   & 3  & 5.875 & 9.697     & 4.611 & 3.720     & 0.040 & 0.238 \\
			& S2S                   & 2  & 8.372 & 15.599     & 5.403 & 6.728      & 0.033 & 0.262  \\
			\cmidrule{1-9} 
			\multicolumn{1}{c}{\multirow{5}[0]{*}{3}} 
			& Vel.                  & 0  & 2.093 & 5.436    & 2.709  & 11.807      & 0.177 & 0.401 \\
			& S2M                   & 3   & 0.368 & 0.225    & 0.468 & 8.769      & 0.149 & 0.343 \\
			& Both                  & 0  & 0.051 & 0.032     & 0.303 & 8.368      & 0.132 & 0.343 \\
			& CVM                   & 3  & 6.183 & 9.613     & 4.049 & 12.607     & 0.198 & 0.639 \\
			& S2S                   & 1  & 3.689 & 1.482     & 2.067 & 10.694      & 0.168 & 0.531  \\
			\bottomrule
		\end{tabular}%
		\begin{tablenotes}
			\item[a] Vel.: radar ego-velocity constraint, S2M: scan-to-submap matching, Both: Vel. and S2M, CVM: constant velocity model for state prediction, S2S: scan-to-five-scans matching.
			\item[b] F means the number of failures in 10 repetitions.
			\item[c] APE Rot. RMSEs are in $^\circ$, APE Trans. RMSEs and Closure Errors are in m. RPE Rot. RMSEs are in $^\circ/m$, and RPE Trans. RMSEs are in $\%$.
		\end{tablenotes}
	\end{threeparttable}
\end{table}%

From Table~\ref{tab:addlabel2}, we see that:

(1) Closure errors of the three sequences are large when only ego-velocity constraints are used, especially in the Z-direction, and the APE RMSE in position exceeds 2 m.

(2) When only scan-to-submap matching is used, estimation failure occurs frequently because of match failures or incorrect matches. 
However, in successful operations, the location accuracy is relatively high and the trajectory closure error is small.
This shows that scan-to-submap matching can mitigate the trajectory drift in the Z-direction especially for small scenes (sequence 1 and 3). 

(3) When the velocity constraints and scan-to-submap matching are both used, no estimation failure occurs in the repeated runs of the three sequences.
The mean closure errors and RMSEs of the trajectories are also largely reduced.

(4) When using the constant velocity model for state prediction, the localization accuracy of the system obviously drops.
Empirically, more than 3 out of 10 times, the algorithm failed to estimate the trajectory completely without IMU data, 
and it often deviated from the real trajectory severely midway.

(5) When the size of the temporal window for scan matching is one, i.e., the exact scan-to-scan method, the algorithm stopped working midway in all repeated runs, 
and the trajectory estimation failed. 
When the temporal window for scan matching has 5 scans,
the algorithm performance significantly improved.
However, since this scan-to-five-scan matching only considered the temporally aggregated points instead of all the local points,
the algorithm accuracy was still lower than that of the scan-to-submap matching used in iRIOM.
Continuing to enlarge the temporal window size greatly increased the computation cost and its benefit on accuracy flattened out.

\subsection{ColoRadar Dataset Results}

\begin{table}[!thpb]
	\centering
	\caption{Statistics of EKFRIO, iRIO, and iRIOM on ColoRadar}
	\label{tab:addtable3}%
	\begin{threeparttable}
		\begin{tabular}{cccccc}
			\toprule
			\multicolumn{1}{c}{\multirow{2}[3]{*}{}}  & \multirow{2}[3]{*}{Method}  & \multicolumn{2}{c}{Closure Error}  & \multicolumn{2}{c}{APE RMSE}  \\
			\cmidrule(r){3-4}      \cmidrule(r){5-6}       
			&       & Hor (m)       & Vert (m)            & Trans (m)       & Rot ($^{\circ} $)    \\
			\cmidrule{1-6} 
			\multicolumn{1}{c}{\multirow{3}[0]{*}{Seq 1}} 
			& EKFRIO           & 15.732    & 6.746      & 5.048  & 10.758     \\
			& iRIO             & 7.309     & 0.652       & 2.377  & 7.896     \\
			& iRIOM            & 0.351     & 0.043       & 0.916  & 7.198    \\
			\cmidrule{1-6} 
			\multicolumn{1}{c}{\multirow{3}[0]{*}{Seq 2}}
			& EKFRIO           & 0.619   & 8.045       & 2.353  & 12.795       \\
			& iRIO             & 1.254   & 0.909       & 0.610  & 9.050        \\
			& iRIOM            & 0.185   & 0.195       & 0.385  & 8.222        \\
			\bottomrule
		\end{tabular}%
		\begin{tablenotes}
			\item[a] {Seq 1: 2\_23\_2021\_edgar\_classroom\_run0.\\ Seq 2: 2\_28\_2021\_outdoors\_run0.}
		\end{tablenotes}
	\end{threeparttable}
\end{table}%

\begin{figure}[thpb]
	\centering
	\includegraphics[width=1\linewidth]{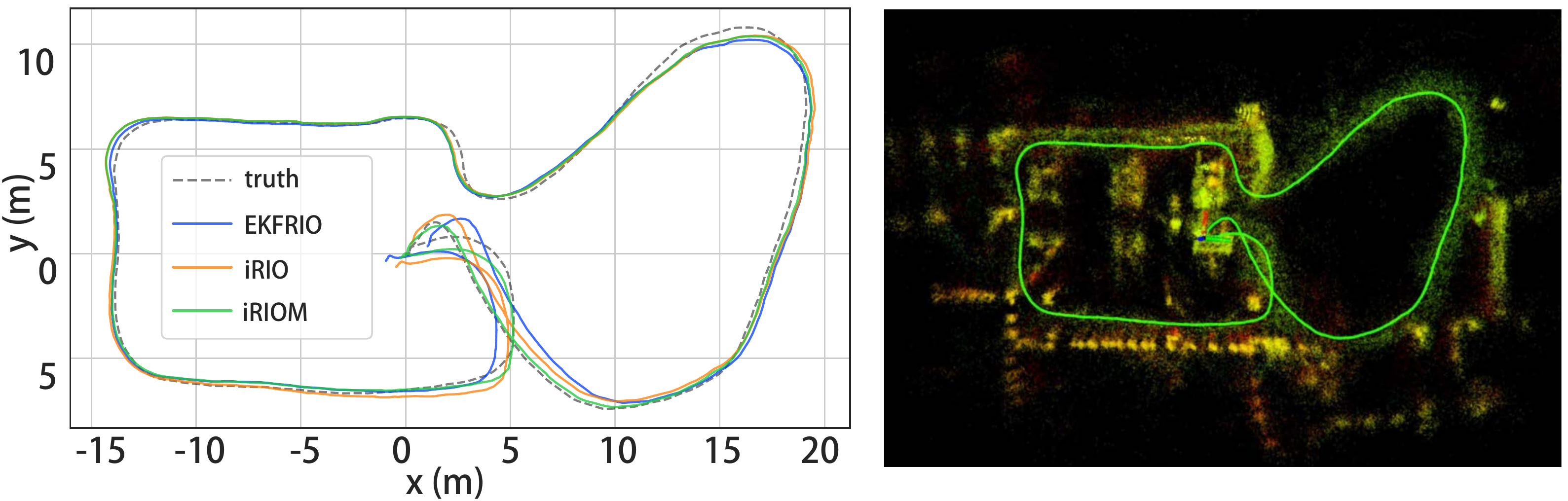}
	\caption{Left: Trajectory estimated by EKFRIO, iRIO, and iRIOM on 2\_28\_2021\_outdoors\_run0 sequence. Right: The mapping result of iRIOM.}
	\label{fig:addfig3_}
\end{figure}

To verify the generality of our algorithm, we ran experiments on the ColoRadar sequences \cite{kramer2022coloradar}. 
As shown in Table~\ref{tab:addtable3} and Fig.~\ref{fig:addfig3_}, the experimental results on indoor (2\_23\_2021\_edgar\_classroom\_run0) and outdoor (2\_28\_2021\_outdoors\_run0) sequences indicate 
that proposed algorithm has good generality while achieving high accuracy localization and mapping.

\section{CONCLUSIONS}

This paper presents a 4D radar inertial odometry and mapping method, iRIOM, in 3D space,
which fuses IMU data, ego-velocity from a radar point cloud scan, and point matches from scan-to-submap matching.
The radar velocity relative to the static world is estimated by the
GNC method, addressing moving objects and multipath effects.
Scan-to-submap matching leverages the accumulative points in a local area, 
and the point matches are weighted by a distribution-to-multi-distribution scheme to deal with non-exact point matching.
Experimental results showed that our method achieved accurate odometry and consistent mapping wrt. EKFVIO and FastLIO-SLAM on our and ColoRadar dataset.
Ablation study validated the benefits of IMU data, scan-to-submap matching, and loop closure.
To our knowledge, iRIOM is the first to achieve real-time 3D odometry and consistent mapping using 4D radar data.

In the future, we would like to explore other constraints, 
such as the Manhattan world assumption \cite{doer2021yaw} and ground planes \cite{wei2022gclo}, and integrate other sensor modalities, e.g., cameras \cite{huai2022observability}.

\section*{Supplementary Material}
A video showing iRIOM on our and ColoRadar sequences is at 
\url{https://youtu.be/Fl9CbSmbd8M}.

\section*{Acknowledgments}
We thank the topical editor and anonymous reviewers for their helpful and inspiring comments. Also, downloading the ColoRadar dataset by Yuxin Shao is deeply appreciated.

%bib
\bibliographystyle{IEEEtran}
\bibliography{ms}

\begin{thebibliography}{10}
\providecommand{\url}[1]{#1}
\csname url@samestyle\endcsname
\providecommand{\newblock}{\relax}
\providecommand{\bibinfo}[2]{#2}
\providecommand{\BIBentrySTDinterwordspacing}{\spaceskip=0pt\relax}
\providecommand{\BIBentryALTinterwordstretchfactor}{4}
\providecommand{\BIBentryALTinterwordspacing}{\spaceskip=\fontdimen2\font plus
\BIBentryALTinterwordstretchfactor\fontdimen3\font minus
  \fontdimen4\font\relax}
\providecommand{\BIBforeignlanguage}[2]{{%
\expandafter\ifx\csname l@#1\endcsname\relax
\typeout{** WARNING: IEEEtran.bst: No hyphenation pattern has been}%
\typeout{** loaded for the language `#1'. Using the pattern for}%
\typeout{** the default language instead.}%
\else
\language=\csname l@#1\endcsname
\fi
#2}}
\providecommand{\BIBdecl}{\relax}
\BIBdecl

\bibitem{dickmann2016automotive}
J.~Dickmann, J.~Klappstein, M.~Hahn, N.~Appenrodt, H.-L. Bloecher, K.~Werber,
  and A.~Sailer, ``Automotive radar the key technology for autonomous driving:
  From detection and ranging to environmental understanding,'' in \emph{2016
  IEEE Radar Conference (RadarConf)}.\hskip 1em plus 0.5em minus 0.4em\relax
  {Philadelphia, PA, USA}: IEEE, 2016, pp. 1--6.

\bibitem{brooker2001millimetre}
G.~Brooker, M.~Bishop, and S.~Scheding, ``Millimetre waves for robotics,'' in
  \emph{Australian Conference for Robotics and Automation}, 2001.

\bibitem{kellner2013instantaneous}
D.~Kellner, M.~Barjenbruch, J.~Klappstein, J.~Dickmann, and K.~Dietmayer,
  ``Instantaneous ego-motion estimation using {D}oppler radar,'' in \emph{16th
  IEEE Int. Conf. on Intelligent Transportation Systems (ITSC 2013)}.\hskip 1em
  plus 0.5em minus 0.4em\relax IEEE, 2013, pp. 869--874.

\bibitem{sun20214d}
S.~Sun and Y.~D. Zhang, ``4d automotive radar sensing for autonomous vehicles:
  A sparsity-oriented approach,'' \emph{IEEE Journal of Selected Topics in
  Signal Processing}, vol.~15, no.~4, pp. 879--891, 2021.

\bibitem{doer2021yaw}
C.~Doer and G.~F. Trommer, ``Yaw aided radar inertial odometry using manhattan
  world assumptions,'' in \emph{2021 28th Saint Petersburg Int. Conf. on
  Integrated Navigation Systems (ICINS)}.\hskip 1em plus 0.5em minus
  0.4em\relax {St. Petersburg, Russia}: IEEE, 2021, pp. 1--9.

\bibitem{doer2020ekf}
------, ``An {EKF} based approach to radar inertial odometry,'' in \emph{2020
  IEEE Int. Conf. on Multisensor Fusion and Integration for Intelligent Systems
  (MFI)}.\hskip 1em plus 0.5em minus 0.4em\relax {Karlsruhe, Germany}: IEEE,
  2020, pp. 152--159.

\bibitem{michalczyktightly}
J.~Michalczyk, R.~Jung, and S.~Weiss, ``Tightly-coupled {EKF}-based
  radar-inertial odometry,'' in \emph{2022 IEEE/RSJ International Conference on
  Intelligent Robots and Systems (IROS)}.\hskip 1em plus 0.5em minus
  0.4em\relax {Kyoto, Japan}: IEEE, 2022, pp. 12\,336--12\,343.

\bibitem{kim2018scan}
G.~Kim and A.~Kim, ``Scan context: Egocentric spatial descriptor for place
  recognition within {3D} point cloud map,'' in \emph{2018 IEEE/RSJ Int. Conf.
  on Intelligent Robots and Systems (IROS)}.\hskip 1em plus 0.5em minus
  0.4em\relax {Madrid, Spain}: IEEE, 2018, pp. 4802--4809.

\bibitem{kramer2022coloradar}
A.~Kramer, K.~Harlow, C.~Williams, and C.~Heckman, ``Coloradar: The direct {3D}
  millimeter wave radar dataset,'' \emph{The International Journal of Robotics
  Research}, vol.~41, no.~4, pp. 351--360, 2022.

\bibitem{cen2018precise}
S.~H. Cen and P.~Newman, ``Precise ego-motion estimation with millimeter-wave
  radar under diverse and challenging conditions,'' in \emph{2018 IEEE Int.
  Conf. on Robotics and Automation (ICRA)}.\hskip 1em plus 0.5em minus
  0.4em\relax {Brisbane, Australia}: IEEE, 2018, pp. 6045--6052.

\bibitem{cen2019radar}
------, ``Radar-only ego-motion estimation in difficult settings via graph
  matching,'' in \emph{2019 Int. Conf. on Robotics and Automation
  (ICRA)}.\hskip 1em plus 0.5em minus 0.4em\relax {Montreal, QC, Canada}: IEEE,
  2019, pp. 298--304.

\bibitem{aldera2019fast}
R.~Aldera, D.~De~Martini, M.~Gadd, and P.~Newman, ``Fast radar motion
  estimation with a learnt focus of attention using weak supervision,'' in
  \emph{2019 Int. Conf. on Robotics and Automation (ICRA)}.\hskip 1em plus
  0.5em minus 0.4em\relax {Montreal, QC, Canada}: IEEE, 2019, pp. 1190--1196.

\bibitem{barnes2020under}
D.~Barnes and I.~Posner, ``Under the radar: Learning to predict robust
  keypoints for odometry estimation and metric localisation in radar,'' in
  \emph{2020 IEEE Int. Conf. on Robotics and Automation (ICRA)}.\hskip 1em plus
  0.5em minus 0.4em\relax {Virtual}: IEEE, 2020, pp. 9484--9490.

\bibitem{hong2020radarslam}
Z.~Hong, Y.~Petillot, and S.~Wang, ``Radarslam: Radar based large-scale {SLAM}
  in all weathers,'' in \emph{2020 IEEE/RSJ Int. Conf. on Intelligent Robots
  and Systems (IROS)}.\hskip 1em plus 0.5em minus 0.4em\relax {Las Vegas, USA}:
  IEEE, 2020, pp. 5164--5170.

\bibitem{burnett2021we}
K.~Burnett, A.~P. Schoellig, and T.~D. Barfoot, ``Do we need to compensate for
  motion distortion and doppler effects in spinning radar navigation?''
  \emph{IEEE Robotics and Automation Letters}, vol.~6, no.~2, pp. 771--778,
  2021.

\bibitem{park2020pharao}
Y.~S. Park, Y.-S. Shin, and A.~Kim, ``Pharao: Direct radar odometry using phase
  correlation,'' in \emph{2020 IEEE Int. Conf. on Robotics and Automation
  (ICRA)}.\hskip 1em plus 0.5em minus 0.4em\relax {Virtual}: IEEE, 2020, pp.
  2617--2623.

\bibitem{barnes2019masking}
D.~Barnes, R.~Weston, and I.~Posner, ``Masking by moving: Learning
  distraction-free radar odometry from pose information,'' \emph{arXiv preprint
  arXiv:1909.03752}, 2019.

\bibitem{almalioglu2020milli}
Y.~Almalioglu, M.~Turan, C.~X. Lu, N.~Trigoni, and A.~Markham, ``Milli-rio:
  Ego-motion estimation with low-cost millimetre-wave radar,'' \emph{IEEE
  Sensors Journal}, vol.~21, no.~3, pp. 3314--3323, 2020.

\bibitem{lu2020milliego}
C.~X. Lu, M.~R.~U. Saputra, P.~Zhao, Y.~Almalioglu, P.~P. De~Gusmao, C.~Chen,
  K.~Sun, N.~Trigoni, and A.~Markham, ``Milliego: Single-chip mmwave radar
  aided egomotion estimation via deep sensor fusion,'' in \emph{Proceedings of
  the 18th Conference on Embedded Networked Sensor Systems}, 2020, pp.
  109--122.

\bibitem{kramer2020radar}
A.~Kramer, C.~Stahoviak, A.~Santamaria-Navarro, A.-A. Agha-Mohammadi, and
  C.~Heckman, ``Radar-inertial ego-velocity estimation for visually degraded
  environments,'' in \emph{2020 IEEE Int. Conf. on Robotics and Automation
  (ICRA)}.\hskip 1em plus 0.5em minus 0.4em\relax {Virtual}: IEEE, 2020, pp.
  5739--5746.

\bibitem{li2021real}
Q.~Li, J.~Huai, D.~Chen, and Y.~Zhuang, ``Real-time robot localization based on
  {2D} lidar scan-to-submap matching,'' in \emph{China Satellite Navigation
  Conference (CSNC 2021) Proceedings}.\hskip 1em plus 0.5em minus 0.4em\relax
  {Beijing, China}: Springer, 2021, pp. 414--423.

\bibitem{wang2021radarloc}
W.~Wang, P.~P. de~Gusmao, B.~Yang, A.~Markham, and N.~Trigoni, ``Radarloc:
  Learning to relocalize in {FMCW} radar,'' in \emph{2021 IEEE Int. Conf. on
  Robotics and Automation (ICRA)}.\hskip 1em plus 0.5em minus 0.4em\relax
  {Xi'an, China}: IEEE, 2021, pp. 5809--5815.

\bibitem{koide2021voxelized}
K.~Koide, M.~Yokozuka, S.~Oishi, and A.~Banno, ``Voxelized {GICP} for fast and
  accurate {3D} point cloud registration,'' in \emph{2021 IEEE Int. Conf. on
  Robotics and Automation (ICRA)}.\hskip 1em plus 0.5em minus 0.4em\relax
  {Xi'an, China}: IEEE, 2021, pp. 11\,054--11\,059.

\bibitem{balta2018fast}
H.~Balta, J.~Velagic, W.~Bosschaerts, G.~De~Cubber, and B.~Siciliano, ``Fast
  statistical outlier removal based method for large {3D} point clouds of
  outdoor environments,'' \emph{IFAC-PapersOnLine}, vol.~51, no.~22, pp.
  348--353, 2018.

\bibitem{yangGraduatedNonconvexityRobust2020}
H.~Yang, P.~Antonante, V.~Tzoumas, and L.~Carlone, ``Graduated non-convexity
  for robust spatial perception: {{From}} non-minimal solvers to global outlier
  rejection,'' \emph{IEEE Robotics and Automation Letters}, vol.~5, no.~2, pp.
  1127--1134, Apr. 2020.

\bibitem{yangCertifiablyOptimalOutlierrobust2022}
H.~Yang and L.~Carlone, ``Certifiably optimal outlier-robust geometric
  perception: {{Semidefinite}} relaxations and scalable global optimization,''
  \emph{IEEE Transactions on Pattern Analysis and Machine Intelligence},
  vol.~45, no.~3, pp. 2816--2834, 2022.

\bibitem{xu2022fast}
W.~Xu, Y.~Cai, D.~He, J.~Lin, and F.~Zhang, ``Fast-{LIO2}: Fast direct
  lidar-inertial odometry,'' \emph{IEEE Transactions on Robotics}, 2022.

\bibitem{huai2022observability}
J.~Huai, Y.~Lin, Y.~Zhuang, C.~K. Toth, and D.~Chen, ``Observability analysis
  and keyframe-based filtering for visual inertial odometry with full
  self-calibration,'' \emph{IEEE Transactions on Robotics}, vol.~38, no.~5, pp.
  3219--3237, 2022.

\bibitem{wei2022gclo}
X.~Wei, J.~Lv, J.~Sun, E.~Dong, and S.~Pu, ``Gclo: Ground constrained lidar
  odometry with low-drifts for {GPS}-denied indoor environments,'' in
  \emph{2022 Int. Conf. on Robotics and Automation (ICRA)}.\hskip 1em plus
  0.5em minus 0.4em\relax {Philadelphia PA, USA}: IEEE, 2022, pp. 2229--2235.

\end{thebibliography}

\vfill

\end{document}